\documentclass{article}

\usepackage[preprint]{neurips_2026}

\usepackage[utf8]{inputenc}
\usepackage[T1]{fontenc}
\usepackage{hyperref}
\usepackage{url}
\usepackage{booktabs}
\usepackage{amsfonts}
\usepackage{amsmath}
\usepackage{amssymb}
\usepackage{nicefrac}
\usepackage{microtype}
\usepackage{xcolor}
\usepackage{colortbl}
\usepackage{multirow}
\usepackage{graphicx}
\usepackage{enumitem}
\usepackage{placeins}
\usepackage{float}
\usepackage{wrapfig}
\usepackage{subcaption}

\makeatletter

\newcommand{\tabstyle}[4]{
    \centering
    \resizebox{#1\textwidth}{!}{
    \setlength\tabcolsep{#2pt}
    \renewcommand\arraystretch{#3}
    #4
    }
}

\makeatother

\definecolor{mygray}{gray}{.95}
\definecolor{lightgray}{gray}{.95}
\definecolor{lightblue}{RGB}{208,235,255}
\definecolor{lightpink}{RGB}{255,222,235}
\definecolor{ggray}{RGB}{127,127,127}
\definecolor{mygreen}{RGB}{93,173,85}
\definecolor{myyellow}{RGB}{190,144,0}
\definecolor{myblue}{RGB}{30,90,100}

\def\ie{\textit{i.e.}}
\def\eg{\textit{e.g.}}

\usepackage{pifont}

\newcommand{\cmark}{\textcolor{green!60!black}{\ding{51}}}
\newcommand{\xmark}{\textcolor{red!75!black}{\ding{55}}}

\usepackage[table]{xcolor}
\usepackage{ulem}
\newcommand{\best}[1]{\textcolor{red}{\textbf{#1}}}
\newcommand{\secondbest}[1]{\textcolor{blue}{\underline{#1}}}

\title{\textsc{Trace}: Temporal Routing with Autoregressive Cross-channel Experts for EEG Representation Learning}

\author{%
  Fan Ma\thanks{Fan Ma and Qier An contributed equally to this work.}
  \quad Qier An\textsuperscript{*}
  \quad Peng Chen
  \quad Lingfei Qian
  \quad Xiang Lan \\
  \bfseries Mingyang Jiang
  \quad Zhiling Gu
  \quad Xenophon Papademetris
  \quad Hua Xu\thanks{Corresponding author.} \\
  \normalfont Department of Biomedical Informatics and Data Science, Yale University
}

\begin{document}

\maketitle

\begin{abstract}
  Learning transferable representations for electroencephalography (EEG) remains challenging because EEG signals are inherently multi-channel and non-stationary. Channels observed at the same time provide coupled measurements of neural activity, while the relevant temporal dynamics vary across contexts.
  This structure is poorly matched by architectures that apply uniform computation across time or route each channel patch independently. To this end, we propose \textsc{Trace}, an autoregressive EEG pre-training framework that predicts future EEG patches from causal context while performing temporally adaptive and cross-channel coherent computation. At each temporal step, \textsc{Trace} derives an expert routing decision from the causal cross-channel history and applies it jointly to all channels at that step. This preserves instantaneous cross-channel coherence while allowing different temporal regimes to activate different computation. Since routing is defined over the available channel set and causal temporal context, \textsc{Trace} is compatible with heterogeneous pre-training across corpora with different channel counts, montages, sequence lengths, and recording domains.
Across eight downstream EEG benchmarks, \textsc{Trace} is evaluated in both settings: when downstream domains are seen only as unlabeled pre-training data and when downstream datasets are completely unseen during pre-training. It obtains the best results on several benchmarks while remaining competitive on motor imagery and clinical event classification tasks, with ablations supporting the importance of cross-channel temporal routing.
\end{abstract}

\section{Introduction}
\label{sec:intro}

Electroencephalography (EEG) is widely used in brain-computer interfaces (BCIs) and clinical monitoring because it is non-invasive and provides high temporal resolution~\citep{lawhern2018eegnet,song2022eeg,ma2025codebrain}. However, learning transferable EEG representations remains challenging: EEG signals are noisy, non-stationary, and highly variable across montages, recording protocols, temporal scales, and subject populations~\citep{jiang2024large,wang2025cbramod,ma2025codebrain}. Consequently, models trained on one dataset or electrode layout often transfer poorly to another, motivating foundation models that learn general-purpose representations from large unlabeled EEG corpora.

Recent EEG foundation models are first pre-trained on large-scale unlabeled EEG and then fine-tuned on downstream tasks. Representative methods, including BIOT~\citep{yang2023biot}, LaBraM~\citep{jiang2024large}, CBraMod~\citep{wang2025cbramod}, and CodeBrain~\citep{ma2025codebrain}, have achieved noticeable improvements over task-specific architectures.
Most follow a BERT-style paradigm~\citep{devlin2018bert}, masking portions of the input and reconstructing missing signals, tokens, or latent codes. Although effective, masked reconstruction is not the only natural objective for learning EEG representation.
EEG signals are inherently directional in time, reflecting neural activity that unfolds as a sequence of evolving states. While effective for static classification, masked modeling may overlook the intrinsic causal dynamics that are crucial for applications such as online continuous clinical monitoring. These applications require representations that not only encode the current brain state but also capture the transition rules to infer future neural dynamics from past observations.

Autoregressive (AR) modeling naturally aligns with this setting by predicting future EEG observations from causal context. Recent studies have begun to explore generative or predictive objectives for EEG pre-training, \eg, NeuroLM~\citep{jiang2024neurolm} and EEGPT~\citep{yue2024eegpt}. However, AR pre-training alone does not address a key architectural challenge in multi-channel EEG: computation should adapt to changing temporal brain states while preserving the coherence among channels observed at the same time.

One natural way to improve model capacity and adaptability is to introduce mixture-of-experts (MoE), which can allocate different computational pathways to different neural states~\citep{shazeer2017outrageously}.
However, directly applying generic token-wise MoE routing is not well suited to multi-channel EEG.
In standard token-wise routing, each channel patch is routed independently.
Such statistical decoupling overlooks the fact that channels recorded at the same temporal step are different observations of a shared latent brain state, shaped by synchronized neural activity.
Assigning these temporally aligned channel tokens to different experts may weaken cross-channel coherence and make it harder for experts to specialize in meaningful neural dynamics.
Thus, we argue that expert specialization should be conditioned on the temporal state of the whole brain rather than on isolated channel patches, enabling adaptive computation across non-stationary neural states while preserving cross-channel consistency.

In this study, we introduce \textsc{Trace}, a \textbf{\underline{T}emporal \underline{R}outing with \underline{A}utoregressive \underline{C}ross-channel \underline{E}xperts} framework for EEG pre-training.
\textsc{Trace} predicts future EEG patches from causal context, but differs from autoregressive Transformers in how it allocates computation across time. Its core Temporal Routing MoE (TR-MoE) block combines spatial-temporal attention with a Cross-Channel Temporal Routing FFN (CTR-FFN). Within CTR-FFN, a TemporalFormer router summarizes the causal cross-channel history, and an Expert Selector uses this temporal-state representation to route all channels at the same temporal step to the same experts. This design preserves cross-channel coherence while adapting computation to non-stationary temporal EEG states.

\textsc{Trace} also supports heterogeneous EEG pre-training across different montages, recording protocols, and sequence lengths without projecting all recordings onto a common montage. Instead of assuming a fixed electrode layout, \textsc{Trace} encodes each available channel as a temporal patch sequence and computes routing from the causal cross-channel state present in each sample. Our pre-training corpus combines clinical EEG, high-density healthy-population EEG, and task-related EEG, covering $16$-$128$ channels and $4$-$30$ second windows. This diversity exposes the model to a broad range of acquisition settings and temporal dynamics while keeping the encoder and CTR-FFN compatible with variable channel counts.

We evaluate \textsc{Trace} on eight downstream datasets spanning six BCI task categories under two transfer regimes: seen-domain transfer, where the downstream domain is observed only through unlabeled pre-training data, and unseen-dataset generalization, where the downstream dataset is completely excluded from pre-training. \textsc{Trace} achieves leading results on several benchmarks and remains competitive on the remaining motor imagery and clinical event classification tasks, while ablations confirm the benefits of CTR-FFNs and multi-source pre-training.
To summarize, our main contributions are four-fold:

\begin{itemize}[leftmargin=1.4em, itemsep=2pt, topsep=3pt]
    \item We propose \textsc{Trace}, an autoregressive EEG pre-training framework that forecasts future EEG patches from causal context while centering the architecture on temporally adaptive and cross-channel coherent computation.
    \item We introduce the TR-MoE block for multi-channel EEG, combining Spatial-Temporal Attention with a CTR-FFN that routes same-step channels via a shared temporal-state decision.
    \item We pre-train \textsc{Trace} on a heterogeneous multi-source EEG corpus spanning different channel counts, acquisition settings, task paradigms, and temporal dynamics, without enforcing the same common montage across all datasets.
    \item We show that \textsc{Trace} achieves state-of-the-art performance on several downstream EEG datasets, and validate our designs through comprehensive ablations.
\end{itemize}

\section{Method}

\begin{figure}[t]
    \centering
    \includegraphics[width=\textwidth]{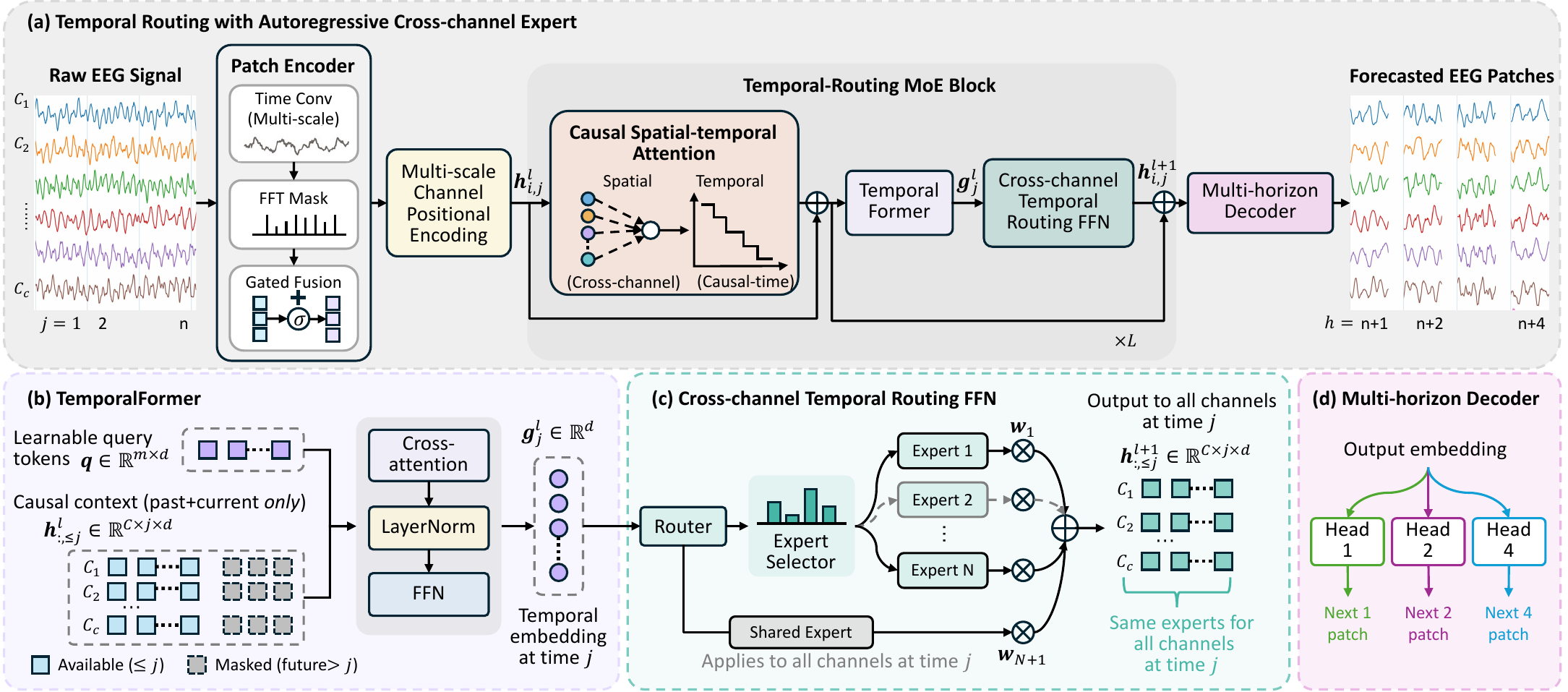}
    \caption{\textbf{Overview of \textsc{Trace}.} \textsc{Trace} partitions raw EEG into channel-wise temporal patches, embeds each patch with time-frequency features and multi-scale channel positional encoding, and processes the resulting sequence using stacked Temporal Routing MoE (TR-MoE) blocks. Each TR-MoE block combines causal spatial-temporal attention with a Cross-Channel Temporal Routing FFN (CTR-FFN). For each temporal step, TemporalFormer summarizes the causal cross-channel context, and a router selects top-$K$ temporal experts shared by all channels at that step, together with an always-on shared expert. A multi-horizon decoder predicts future EEG patches under autoregressive and expert-balancing objectives.}
    \label{fig:trace_framework}
\end{figure}

\subsection{Problem Setup and Causal Patch Representation}
\label{sec:method:setup}

Given an EEG segment $\mathbf{S} \in \mathbb{R}^{C \times T}$ with $C$ channels and $T$ timestamps, we split each channel into $n=\lfloor T/t \rfloor$ non-overlapping patches of length $t$:
\begin{equation}
    \mathbf{X}
    =
    \{\mathbf{x}_{i,j} \mid i=1,\ldots,C,\; j=1,\ldots,n\},
    \qquad
    \mathbf{x}_{i,j} \in \mathbb{R}^{t}
\end{equation}
where $i$ indexes channels and $j$ indexes temporal patches. \textsc{Trace} learns a causal patch representation: the representation at temporal step $j$ is computed only from patches up to $j$, \ie, it cannot access $\{\mathbf{x}_{i,j'} \mid j'>j\}$. The learned representation is then used to predict future patches autoregressively.

\subsection{Overview of \textsc{Trace}}
\label{sec:method:overview}

As shown in Figure~\ref{fig:trace_framework}, \textsc{Trace} first maps raw channel-wise patches into time-frequency embeddings with a patch encoder and channel positional encoding. A causal transformer backbone then processes the patch sequence while preserving the autoregressive constraint. The core component of the backbone is a Temporal Routing MoE (TR-MoE) block, where a Cross-Channel Temporal Routing FFN (CTR-FFN) uses TemporalFormer to select temporal experts shared across channels at each step. Finally, \textsc{Trace} is pre-trained with an autoregressive multi-horizon objective that predicts multiple future patch ranges from each causal representation.

\subsection{Patch Encoder and Positional Encoding}
\label{sec:method:encoder}
The patch encoder maps each channel-wise temporal patch into a $d$-dimensional time-frequency embedding. It consists of three modules: in-patch multi-scale convolution for time-domain features, FFT-based spectral masking for frequency-domain features, and gated fusion for combining the two. Throughout this section, a \emph{temporal patch} refers to the patch $\mathbf{x}_{i,j}$ of length $t$ defined in Section~\ref{sec:method:setup}.

\paragraph{In-Patch Multi-scale Convolution.}
The first module applies several 1D convolutional branches with different temporal kernel sizes to each patch, allowing the encoder to capture local waveform patterns at multiple scales within the same patch. Let $\mathcal{Q}_{\mathrm{temp}}$ denote the set of temporal kernel sizes. We summarize this module as:

\begin{align}
    \phi_{q}(\mathbf{x}_{i,j})
    &=
    \operatorname{GELU}\!\left(
        \operatorname{GN}\!\left(
            \operatorname{Conv1d}_{3}
            \left(
                \operatorname{GELU}\!\left(
                    \operatorname{GN}\!\left(
                        \operatorname{Conv1d}_{q}(\mathbf{x}_{i,j})
                    \right)
                \right)
            \right)
        \right)
    \right),
    \qquad q \in \mathcal{Q}_{\mathrm{temp}}, \\
    \mathbf{e}^{\mathrm{temp}}_{i,j}
    &=
    \mathbf{W}_{\mathrm{temp}}
    \left[
        \phi_{q}(\mathbf{x}_{i,j})
    \right]_{q \in \mathcal{Q}_{\mathrm{temp}}}
    \in \mathbb{R}^{d}.
\end{align}
where $\phi_q(\cdot)$ is the branch output for kernel size $q$, $[\cdot]_{q \in \mathcal{Q}_{\mathrm{temp}}}$ denotes concatenation over kernel sizes, and $\operatorname{GN}(\cdot)$ denotes group normalization.

\paragraph{FFT-Based Spectral Masking.}
The second module transforms the same patch into the frequency domain and applies a learnable spectral mask to emphasize informative frequency components before projection by:

\begin{equation}
    \mathbf{e}^{\mathrm{freq}}_{i,j}
    =
    \operatorname{FC}\!\left(
        \left|
        \mathbf{M}_{\theta}
        \odot
        \mathcal{F}(\mathbf{x}_{i,j})
        \right|
    \right)
    \in \mathbb{R}^{d},
\end{equation}
where $\mathcal{F}(\cdot)$ is the discrete Fourier transform, $\mathbf{M}_{\theta}$ is a learnable spectral mask, and $\odot$ denotes element-wise multiplication.

\paragraph{Gated Fusion.}
The third module fuses the time-domain embedding $\mathbf{e}^{\mathrm{temp}}_{i,j}$ and the frequency-domain embedding $\mathbf{e}^{\mathrm{freq}}_{i,j}$ with a learned gate via:

\begin{align}
    \mathbf{z}_{i,j}
    &=
    \sigma\!\left(
        \mathbf{W}_{\mathrm{g}}
        [\mathbf{e}^{\mathrm{temp}}_{i,j};\mathbf{e}^{\mathrm{freq}}_{i,j}]
    \right), \\
    \mathbf{e}_{i,j}
    &=
    \mathbf{e}^{\mathrm{temp}}_{i,j}
    +
    \mathbf{z}_{i,j} \odot \mathbf{e}^{\mathrm{freq}}_{i,j}.
\end{align}
Thus, $\mathbf{e}_{i,j}$ is the time-frequency embedding of patch $\mathbf{x}_{i,j}$, and the resulting tensor is $\mathbf{E}\in\mathbb{R}^{C\times n\times d}$.

\paragraph{Multi-scale Channel Positional Encoding (MS-ChPE).}
To encode electrode structure without violating causality, we add a multi-scale channel positional encoding (MS-ChPE). Heterogeneous EEG pre-training involves recordings with different channel counts, montages, and electrode layouts. Without an explicit channel positional signal, the model only observes a set of patch embeddings and cannot reliably infer which electrode or spatial location each token corresponds to. MS-ChPE injects channel identity and layout information from the native channel ordering, enabling the model to exploit electrode structure across heterogeneous montages. Inspired by conditional positional encoding~\citep{chu2021conditional, wang2025cbramod}, MS-ChPE uses depth-wise 1D convolutions:
\begin{align}
    \mathbf{E}^{\mathrm{p}}
    &=
    \sum_{r=1}^{R}
    \operatorname{DWConv1d}_{q_r}(\mathbf{E}), \\
    \mathbf{E}^{\mathrm{o}}
    &=
    \mathbf{E} + \mathbf{E}^{\mathrm{p}}.
\end{align}
The convolution is applied along the channel axis at each patch step, so the positional encoding at step $j$ only uses same-step channel embeddings. We set $\mathbf{h}^{0}_{i,j}=\mathbf{e}^{\mathrm{o}}_{i,j}$ as the input to the backbone.

\subsection{Temporal Routing MoE}
\label{sec:method:trmoe}
The Temporal Routing MoE (TR-MoE) block first applies causal spatial-temporal attention (CSTA), then replaces the standard feed-forward sublayer with a Cross-Channel Temporal Routing FFN (CTR-FFN):
\begin{align}
    \mathbf{u}^{l}_{i,j}
    &=
    \operatorname{CSTA}\!\left(
        \operatorname{LN}(\mathbf{h}^{l}_{i,j})
    \right)
    +
    \mathbf{h}^{l}_{i,j}, \\
    \mathbf{h}^{l+1}_{i,j}
    &=
    \operatorname{CTRFFN}\!\left(
        \operatorname{LN}(\mathbf{u}^{l}_{i,j}),\,
        \mathbf{g}^{l}_{j}
    \right)
    +
    \mathbf{u}^{l}_{i,j}.
\end{align}
where $\operatorname{CSTA}(\cdot)$ denotes causal spatial-temporal multi-head attention, which combines cross-channel attention within the same temporal step with causal temporal attention along the patch axis; implementation details are provided in Appendix~\ref{app:implementation_details}. Cross-channel attention supports heterogeneous EEG inputs because each temporal step can be treated as a set of channel tokens: regardless of the dataset-specific channel count or montage, CSTA can model instantaneous inter-channel interactions over the available channels. $\operatorname{CTRFFN}(\cdot)$ denotes the CTR-FFN, and $\mathbf{g}^{l}_{j}\in\mathbb{R}^{d}$ is a temporal routing context produced by TemporalFormer.

\paragraph{Cross-channel Temporal Routing FFN (CTR-FFN).}
It contains $N$ specialized experts and one shared expert. For each temporal step $j$, a router maps $\mathbf{g}^{l}_{j}$ to top-$K$ specialized experts. The selected experts are shared across all channels at that temporal step, so channels from different montages are routed according to their shared temporal context rather than independent channel-specific decisions. This channel-shared routing preserves within-timestep coherence and further supports heterogeneous EEG pre-training with variable channel counts:
\begin{align}
    \operatorname{CTRFFN}(\mathbf{x}, \mathbf{g}^{l}_{j})
    &=
    \sum_{k=1}^{N}
    r_k(\mathbf{g}^{l}_{j})\operatorname{Expert}_{k}(\mathbf{x})
    +
    \operatorname{Expert}_{N+1}(\mathbf{x}), \\
    r_k(\mathbf{g}^{l}_{j})
    &=
    \begin{cases}
        \dfrac{
            \exp(\mathbf{w}_{k}^{\top}\mathbf{g}^{l}_{j})
        }{
            \sum_{k' \in \mathcal{T}_{j}^{l}}
            \exp(\mathbf{w}_{k'}^{\top}\mathbf{g}^{l}_{j})
        },
        & k \in \mathcal{T}_{j}^{l}, \\[6pt]
        0, & \text{otherwise},
    \end{cases}
\end{align}
where $\mathcal{T}_{j}^{l}$ is the top-$K$ expert set selected at layer $l$ and step $j$. The shared expert $\operatorname{Expert}_{N+1}$ is always active and does not participate in the top-$K$ competition, providing common computation across datasets and channel layouts.

\paragraph{TemporalFormer.}
\label{sec:method:timeformer}
TemporalFormer constructs the routing context $\mathbf{g}^{l}_{j}$ from the causal cross-channel history. It uses $m$ learnable query tokens $\mathbf{Q}\in\mathbb{R}^{m\times d}$ that attend to all channel patches up to step $j$:
\begin{equation}
    \tilde{\mathbf{G}}^{l}_{j}
    =
    \operatorname{MSA}\!\left(
        \mathbf{Q},\,
        \mathbf{H}^{l}_{:,\leq j},\,
        \mathbf{H}^{l}_{:,\leq j}
    \right),
    \qquad
    \mathbf{H}^{l}_{:,\leq j}
    =
    \{\mathbf{h}^{l}_{i,j'} \mid i=1,\ldots,C,\; j'\leq j\}.
\end{equation}
The query outputs are pooled and passed through a feed-forward layer:

\begin{equation}
    \mathbf{g}^{l}_{j}
    =
    \operatorname{FFN}\!\left(
        \operatorname{LN}\!\left(
            \operatorname{Pool}(\tilde{\mathbf{G}}^{l}_{j})
        \right)
    \right).
\end{equation}
Since keys and values are restricted to $\mathbf{H}^{l}_{:,\leq j}$, the routing decision at step $j$ is causal. The fixed number of query tokens also lets the router summarize variable-length and variable-channel EEG contexts into a fixed-dimensional routing vector.

\subsection{Autoregressive Multi-horizon Pre-training}
\label{sec:method:objective}

\textsc{Trace} is trained with a multi-horizon autoregressive forecasting loss and an auxiliary expert-balancing loss.

\paragraph{Multi-horizon Forecasting.}
At each channel $i$ and temporal step $j$, the final hidden state $\mathbf{h}^{L}_{i,j}$ summarizes the causal context up to patch $j$. Rather than predicting only the immediate next patch, \textsc{Trace} predicts multiple future ranges from this same representation. Specifically, for each horizon $\rho\in\mathcal{H}$, a prediction head outputs the next $\rho$ patches:
\begin{equation}
    \hat{\mathbf{x}}^{(\rho)}_{i,j+1:j+\rho}
    =
    \operatorname{Head}_{\rho}(\mathbf{h}^{L}_{i,j}).
\end{equation}
For example, when $\mathcal{H}=\{1,2,4\}$, the model learns to predict the next $1$ patch, the next $2$ patches, and the next $4$ patches in parallel. This multi-horizon design encourages the shared backbone to capture both short-range waveform continuity and longer-range EEG dynamics. Because raw EEG is non-stationary and often contains transient artifacts, we use the Huber loss~\citep{huber1992robust, wen2019robusttad}. The autoregressive loss is thus defined as:

\begin{equation}
    \mathcal{L}_{\mathrm{AR}}
    =
    \frac{1}{|\mathcal{H}|}
    \sum_{\rho \in \mathcal{H}}
    \frac{1}{|\mathcal{V}_{\rho}|}
    \sum_{(i,j) \in \mathcal{V}_{\rho}}
    \mathcal{L}_{\delta}\!\left(
        \mathbf{x}_{i,j+1:j+\rho},\,
        \hat{\mathbf{x}}^{(\rho)}_{i,j+1:j+\rho}
    \right),
\end{equation}
where $\mathcal{L}_{\delta}$ denotes the Huber loss with threshold $\delta$, and $\mathcal{V}_{\rho}=\{(i,j)\mid j+\rho\leq n\}$ is the set of valid prediction positions for horizon $\rho$.

\paragraph{Expert Balancing.}
To prevent routing collapse, we use a load-balancing objective adapted from MoE training~\citep{dai2024deepseekmoe}. Let $f_k$ be the fraction of temporal steps routed to expert $k$ within a batch, and let $p_k$ be the average routing probability assigned to expert $k$. The auxiliary loss is:
\begin{equation}
    \mathcal{L}_{\mathrm{aux}}
    =
    N
    \sum_{k=1}^{N}
    f_k p_k .
\end{equation}

\paragraph{Total Objective.}
The final pre-training objective is thus formulated as:

\begin{equation}
    \mathcal{L}
    =
    \mathcal{L}_{\mathrm{AR}}
    +
    \lambda_{\mathrm{aux}}\mathcal{L}_{\mathrm{aux}},
\end{equation}
where $\lambda_{\mathrm{aux}}$ controls the strength of the expert balancing regularization. We set $\lambda_{\mathrm{aux}}{=}1\mathrm{e}{-}2$ in all pre-training experiments, as listed in Table~\ref{tab:pretrain_hparams}.

\providecolor{lightblue}{RGB}{221,235,247}
\providecommand{\cmark}{\ensuremath{\checkmark}}
\providecommand{\xmark}{\textcolor{black!55}{\ensuremath{\times}}}

\section{Experiment}
\label{sec:experiments}

\subsection{Experimental Setup}
\label{sec:exp:setup}

\paragraph{Pre-training Data and Implementation Details.}
We pre-train \textsc{Trace} on a heterogeneous multi-source EEG corpus consisting of the Temple University Hospital EEG Corpus (TUEG)~\citep{obeid2016temple}, HBN~\citep{alexander2017open}, and several task-related datasets, with channel counts ranging from $16$ to $128$. Following prior EEG foundation model preprocessing protocols~\citep{jiang2024large,wang2025cbramod,ma2025codebrain}, EEG signals are band-pass filtered between $0.5$-$75$ Hz, notch-filtered at $60$ Hz, resampled to $200$ Hz, and divided into non-overlapping windows. After preprocessing, the corpus yields over $1.5$ million EEG segments for pre-training.
For downstream datasets that overlap with task-related pre-training sources, only the training-split subjects are used as unlabeled pre-training data, while validation and test subjects are strictly held out. This creates two evaluation regimes: \textit{seen-domain transfer}, where the downstream domain is observed only through unlabeled pre-training data, and \textit{unseen-dataset generalization}, where the downstream dataset is completely excluded from pre-training. \textsc{Trace} is pre-trained with a multi-horizon autoregressive forecasting objective using $\mathcal{H}=\{1,2,4\}$ on $4$ NVIDIA H100 GPUs. Additional training details are provided in Appendix~\ref{app:pretrain_details}.

\paragraph{Downstream Tasks and Datasets.}
We assess the generalization ability of \textsc{Trace} across a diverse set of BCI tasks, including \textit{Sleep Staging} (\eg, ISRUC~\citep{khalighi2016isruc}), \textit{Emotion Recognition} (\eg, SEED-V~\citep{liu2021comparing}, FACED~\citep{zhang2024gnn4eeg}), \textit{Motor Imagery} (\eg, PhysioNet-MI~\citep{schalk2004bci2000}, SHU-MI~\citep{ma2022large}), \textit{Seizure Detection} (\eg, CHB-MIT~\citep{shoeb2009application}), \textit{Imagined Speech} (\eg, BCIC2020-3~\citep{jeong20222020}), and \textit{Event Classification} (\eg, TUEV~\citep{obeid2016temple}). Among the six main downstream benchmarks, CHB-MIT and PhysioNet-MI evaluate seen-domain transfer, whereas FACED, SEED-V, ISRUC, and BCIC2020-3 evaluate unseen-dataset generalization. Detailed dataset information is provided in Appendix~\ref{sec:dataset_details}.

\paragraph{Baselines and Evaluation Metrics.}
We compare \textsc{Trace} against several state-of-the-art methods, grouped into two categories:
\textit{1) Task-specific models}, including EEGNet~\citep{lawhern2018eegnet} and EEGConformer~\citep{song2022eeg}, and
\textit{2) EEG foundation models}, such as BIOT~\citep{yang2023biot}, LaBraM~\citep{jiang2024large}, CBraMod~\citep{wang2025cbramod}, and CodeBrain~\citep{ma2025codebrain}.
For evaluation, we report Balanced Accuracy (B-Acc), area under the precision-recall curve (AUC-PR), and area under the receiver operating characteristic curve (AUROC) on binary tasks, and B-Acc with weighted F1 (F1-W) on multi-class tasks. All results are averaged over five runs with different random seeds.

\subsection{Comparison with State-of-the-Art Methods}
\label{sec:exp:sota}

Table~\ref{tab:sota_comparison} presents comparisons of \textsc{Trace} against several top-leading task-specific and EEG foundation models across six main downstream EEG datasets, additional results on SHU-MI and TUEV are reported in Appendix~\ref{app:more_downstream_results}. All results are reported as mean $\pm$ standard deviation over five random seeds.
As observed, these results demonstrates strong transfer performance, with clear gains on several benchmarks and competitive results on the remaining tasks, leading to two main conclusions:

\begin{table}[t]
\centering
\caption{Performance comparison of various models across six main EEG datasets. The best and second-best results are highlighted in \textcolor{red}{\textbf{red}} and \textcolor{blue}{\underline{blue}}, respectively.}

\label{tab:sota_comparison}
\tabstyle{1.00}{3}{1.2}{
\begin{tabular}{l|ccc|ccc}
\toprule
\multirow{2}{*}{\textbf{Method}}
& \multicolumn{3}{c|}{\textbf{FACED (9-Class)}}
& \multicolumn{3}{c}{\textbf{SEED-V (5-Class)}} \\
\cmidrule(lr){2-4} \cmidrule(lr){5-7}
& Balanced Acc & Cohen's Kappa & Weighted F1
& Balanced Acc & Cohen's Kappa & Weighted F1 \\
\midrule

EEGNet & 0.4090 $\pm$ 0.0122 & 0.3342 $\pm$ 0.0251 & 0.4124 $\pm$ 0.0141

& 0.2961 $\pm$ 0.0102 & 0.1006 $\pm$ 0.0143 & 0.2749 $\pm$ 0.0098 \\

EEGConformer & 0.4559 $\pm$ 0.0125 & 0.3858 $\pm$ 0.0186 & 0.4514 $\pm$ 0.0107

& 0.3537 $\pm$ 0.0112 & 0.1772 $\pm$ 0.0174 & 0.3487 $\pm$ 0.0136 \\

CNN-Transformer & 0.4697 $\pm$ 0.0132 & 0.4017 $\pm$ 0.0168 & 0.4720 $\pm$ 0.0125

& 0.3678 $\pm$ 0.0078 & 0.2072 $\pm$ 0.0183 & 0.3642 $\pm$ 0.0088 \\

FFCL & 0.4673 $\pm$ 0.0158 & 0.3987 $\pm$ 0.0383 & 0.4699 $\pm$ 0.0145

& 0.3641 $\pm$ 0.0092 & 0.2078 $\pm$ 0.0201 & 0.3645 $\pm$ 0.0132 \\

ST-Transformer & 0.4810 $\pm$ 0.0079 & 0.4137 $\pm$ 0.0133 & 0.4795 $\pm$ 0.0096

& 0.3052 $\pm$ 0.0072 & 0.1083 $\pm$ 0.0121 & 0.2833 $\pm$ 0.0105 \\

\midrule

BIOT & 0.5118 $\pm$ 0.0118 & 0.4476 $\pm$ 0.0254 & 0.5136 $\pm$ 0.0112

& 0.3837 $\pm$ 0.0187 & 0.2261 $\pm$ 0.0262 & 0.3856 $\pm$ 0.0203 \\

LaBraM-Base & 0.5273 $\pm$ 0.0107 & 0.4698 $\pm$ 0.0188 & 0.5288 $\pm$ 0.0102

& 0.3976 $\pm$ 0.0138 & 0.2386 $\pm$ 0.0209 & 0.3974 $\pm$ 0.0111 \\

CBraMod & 0.5509 $\pm$ 0.0089 & 0.5041 $\pm$ 0.0122 & 0.5618 $\pm$ 0.0093

& 0.4091 $\pm$ 0.0097 & 0.2569 $\pm$ 0.0143 & 0.4101 $\pm$ 0.0108 \\

CodeBrain

& \secondbest{0.5941} $\pm$ 0.0098

& \secondbest{0.5406} $\pm$ 0.0084

& \secondbest{0.5953} $\pm$ 0.0113

& \secondbest{0.4137} $\pm$ 0.0023

& \secondbest{0.2735} $\pm$ 0.0032

& \secondbest{0.4235} $\pm$ 0.0022 \\

\midrule

\textsc{Trace} (Ours)

& \best{0.6352} $\pm$ 0.0037

& \best{0.5883} $\pm$ 0.0042

& \best{0.6390} $\pm$ 0.0043

& \best{0.4186} $\pm$ 0.0032

& \best{0.2759} $\pm$ 0.0030

& \best{0.4253} $\pm$ 0.0024 \\

\midrule

\multirow{2}{*}{\textbf{Method}}
& \multicolumn{3}{c|}{\textbf{ISRUC (5-Class)}}
& \multicolumn{3}{c}{\textbf{BCIC2020-3 (5-Class)}} \\
\cmidrule(lr){2-4} \cmidrule(lr){5-7}
& Balanced Acc & Cohen's Kappa & Weighted F1
& Balanced Acc & Cohen's Kappa & Weighted F1 \\
\midrule

EEGNet & 0.7154 $\pm$ 0.0121 & 0.7040 $\pm$ 0.0173 & 0.7513 $\pm$ 0.0124

& 0.4413 $\pm$ 0.0096 & 0.3016 $\pm$ 0.0123 & 0.4413 $\pm$ 0.0102 \\

EEGConformer & 0.7400 $\pm$ 0.0133 & 0.7143 $\pm$ 0.0162 & 0.7634 $\pm$ 0.0151

& 0.4506 $\pm$ 0.0133 & 0.3133 $\pm$ 0.0183 & 0.4488 $\pm$ 0.0154 \\

CNN-Transformer & 0.7363 $\pm$ 0.0087 & 0.7129 $\pm$ 0.0121 & 0.7719 $\pm$ 0.0105

& 0.4533 $\pm$ 0.0092 & 0.3166 $\pm$ 0.0118 & 0.4506 $\pm$ 0.0127 \\

FFCL & 0.7277 $\pm$ 0.0182 & 0.7016 $\pm$ 0.0291 & 0.7614 $\pm$ 0.0197

& 0.4678 $\pm$ 0.0197 & 0.3301 $\pm$ 0.0359 & 0.4689 $\pm$ 0.0205 \\

ST-Transformer & 0.7381 $\pm$ 0.0205 & 0.7013 $\pm$ 0.0352 & 0.7681 $\pm$ 0.0175

& 0.4126 $\pm$ 0.0122 & 0.2941 $\pm$ 0.0159 & 0.4247 $\pm$ 0.0138 \\

\midrule

						BIOT & 0.7527 $\pm$ 0.0121& 0.7192 $\pm$ 0.0231& 0.7790 $\pm$ 0.0146&  0.4920 $\pm$ 0.0086& 0.3650 $\pm$ 0.0176&  0.4917 $\pm$ 0.0079\\
						LaBraM & 0.7633 $\pm$ 0.0102& 0.7231 $\pm$ 0.0182&0.7810 $\pm$ 0.0133&0.5060 $\pm$ 0.0155&0.3800 $\pm$ 0.0242&0.5054 $\pm$ 0.0205\\

CBraMod

& \secondbest{0.7865} $\pm$ 0.0110

& 0.7442 $\pm$ 0.0152

& 0.8011 $\pm$ 0.0099

& 0.5373 $\pm$ 0.0108

& 0.4216 $\pm$ 0.0163

& 0.5383 $\pm$ 0.0096 \\

CodeBrain

& 0.7856 $\pm$ 0.0031

& \secondbest{0.7671} $\pm$ 0.0091

& \secondbest{0.8202} $\pm$ 0.0071

& \secondbest{0.6101} $\pm$ 0.0052

& \secondbest{0.5127} $\pm$ 0.0065

& \secondbest{0.6101} $\pm$ 0.0053 \\

\midrule

\textsc{Trace} (Ours)

& \best{0.7997} $\pm$ 0.0013

& \best{0.7711} $\pm$ 0.0048

& \best{0.8209} $\pm$ 0.0040

& \best{0.6315} $\pm$ 0.0055

& \best{0.5393} $\pm$ 0.0069

& \best{0.6316} $\pm$ 0.0057 \\

\midrule
\multirow{2}{*}{\textbf{Method}}
& \multicolumn{3}{c|}{\textbf{CHB-MIT (2-Class)}}
& \multicolumn{3}{c}{\textbf{PhysioNet-MI (4-Class)}} \\
\cmidrule(lr){2-4} \cmidrule(lr){5-7}
& Balanced Acc & AUC-PR & AUROC
& Balanced Acc & Cohen's Kappa & Weighted F1 \\
\midrule

EEGNet &0.5658 $\pm$ 0.0106&0.1914 $\pm$ 0.0182 & 0.8048 $\pm$ 0.0136 &
0.5814 $\pm$ 0.0125&0.4468 $\pm$ 0.0199&0.5796 $\pm$ 0.0115\\
						EEGConformer & 0.5976 $\pm$ 0.0141& 0.2209 $\pm$ 0.0215&0.8226 $\pm$ 0.0170& 0.6049 $\pm$ 0.0104& 0.4736 $\pm$ 0.0171& 0.6062 $\pm$ 0.0095\\

						CNN-Transformer &0.6389 $\pm$ 0.0067 & 0.2479 $\pm$ 0.0227 & 0.8662 $\pm$ 0.0082& 0.6053 $\pm$ 0.0118& 0.4725 $\pm$ 0.0223& 0.6041 $\pm$ 0.0105\\
						FFCL  &  0.6262 $\pm$ 0.0104 & 0.2049 $\pm$ 0.0346 & 0.8271 $\pm$ 0.0051& 0.5726 $\pm$ 0.0092& 0.4323 $\pm$ 0.0182& 0.5701 $\pm$ 0.0079 \\
						ST-Transformer & 0.5915 $\pm$ 0.0195 & 0.1422 $\pm$ 0.0094 & 0.8237 $\pm$ 0.0491& 0.6035 $\pm$ 0.0081& 0.4712 $\pm$ 0.0199&0.6053 $\pm$ 0.0075\\
						\midrule
						BIOT &0.7068 $\pm$ 0.0457 & 0.3277 $\pm$ 0.0460 & 0.8761 $\pm$ 0.0284& 0.6153 $\pm$ 0.0154&0.4875 $\pm$ 0.0272&0.6158 $\pm$ 0.0197\\
						LaBraM & 0.7075 $\pm$ 0.0358&0.3287 $\pm$ 0.0402&0.8679 $\pm$ 0.0199&0.6173 $\pm$ 0.0122&0.4912 $\pm$ 0.0192&0.6177 $\pm$ 0.0141\\

CBraMod

& \secondbest{0.7398} $\pm$ 0.0284

& 0.3689 $\pm$ 0.0382

& 0.8892 $\pm$ 0.0154

& \best{0.6417} $\pm$ 0.0091

& \best{0.5222} $\pm$ 0.0169

& \best{0.6427} $\pm$ 0.0100 \\

CodeBrain

& 0.7273 $\pm$ 0.0240

& \secondbest{0.4377} $\pm$ 0.0288

& \secondbest{0.8961} $\pm$ 0.0174

& -- & -- & -- \\
        \midrule

\textsc{Trace} (Ours)

& \best{0.7453} $\pm$ 0.0178

& \best{0.5921} $\pm$ 0.0795

& \best{0.9220} $\pm$ 0.0232

& \secondbest{0.6342} $\pm$ 0.0090

& \secondbest{0.5122} $\pm$ 0.0120

& \secondbest{0.6365} $\pm$ 0.0096 \\

\bottomrule
\end{tabular}
}
\end{table}

\textbf{1) \textsc{Trace} achieves superior performance on several evaluated datasets.}
The model demonstrates notable improvements on challenging decoding benchmarks.
To be specific, on the imagined speech task, BCIC2020-3, \textsc{Trace} surpasses CodeBrain and CBraMod in Balanced Accuracy by $+2.14\%$ and $+9.42\%$, respectively. Similar gains are observed on the emotion recognition task (FACED), where \textsc{Trace} improves Balanced Accuracy over CodeBrain and CBraMod by $+4.11\%$ and $+8.43\%$. \textsc{Trace} also attains the best performance on SEED-V, ISRUC, and CHB-MIT in Table~\ref{tab:sota_comparison}. These numerical results support the effectiveness of \textsc{Trace} on several representative EEG transfer settings.

\textbf{2) \textsc{Trace} generalizes across diverse EEG tasks and recording settings.}
The performance gains are not limited to a single task domain or to target-dataset exposure. \textsc{Trace} performs well in both evaluation regimes: it achieves strong transfer when the downstream domain is seen only as unlabeled pre-training data, such as CHB-MIT and PhysioNet-MI, and it also improves on several datasets that are completely unseen during pre-training, including FACED, SEED-V, ISRUC, and BCIC2020-3. Meanwhile, the results are not uniformly state-of-the-art: on PhysioNet-MI, \textsc{Trace} is second-best behind CBraMod. This pattern suggests that \textsc{Trace} learns transferable representations across different channel configurations, recording domains, and task protocols, while leaving room for further improvement on some motor imagery and clinical event classification benchmarks.

\subsection{Ablation Study}
\label{sec:exp:ablation}

\paragraph{Effect of TR-MoE.}
\label{para:ctrffn}
We first validate the core architectural design of \textsc{Trace}. For a fair comparison, we construct two controlled variants: 1) a Non-MoE baseline where the CTR-FFN is replaced with a FLOPs-matched standard FFN, and 2) a Vanilla MoE version that keeps the expert structure but allows each channel patch to select experts via a token-wise router.

\begin{wrapfigure}{r}{0.50\textwidth}
  \centering
  \vspace{-8pt}
  \includegraphics[width=\linewidth]{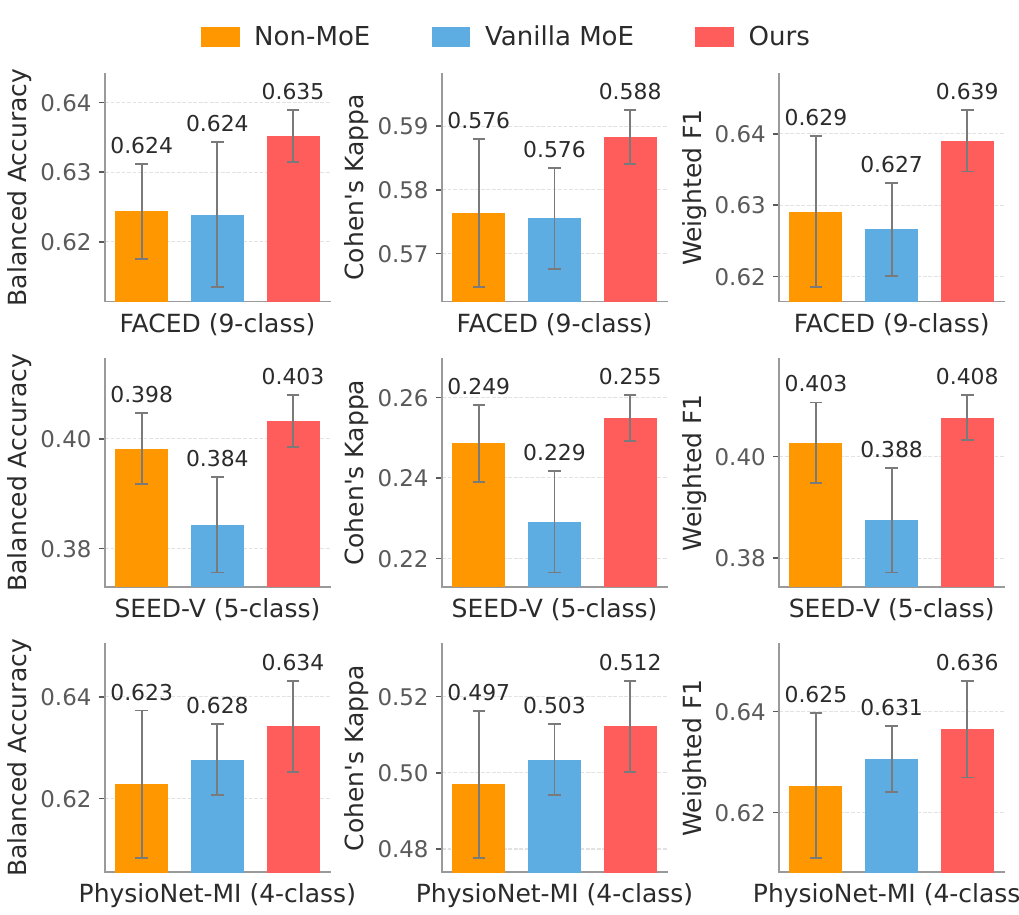}
  \caption{Performance comparison of MoE blocks across three EEG datasets.}
  \label{fig:temporal_moe}
  \vspace{-10pt}
\end{wrapfigure}

Figure~\ref{fig:temporal_moe} shows two key observations.
\textit{First}, standard token-wise MoE does not consistently improve performance: Vanilla MoE is nearly tied with Non-MoE on FACED and even decreases Balanced Accuracy from $0.398$ to $0.384$ on SEED-V, suggesting that simply replacing the FFN with token-wise MoE is insufficient for EEG modeling.
\textit{Second}, the model with TR-MoE blocks achieves the impressive performance across three datasets. Taking FACED as an example, the model with TR-MoE blocks improves over Vanilla MoE by $+1.13\%$ in Balanced Accuracy, $+1.28\%$ in Cohen's Kappa, and $+1.24\%$ in Weighted F1.
This supports our hypothesis that treating all channels at the same timestep as a coherent unit for expert routing is beneficial for capturing cross-channel coordination and temporal EEG dynamics, suggesting that both the MoE structure and temporal-aware routing contribute to the effectiveness of \textsc{Trace}.

\begin{table}[!t]
\centering

\caption{Effect of TemporalFormer query number.
Results are reported as mean $\pm$ std over $5$ random seeds, with the best result in each row highlighted in \textbf{bold}.}

\label{tab:temporalformer_query}
\footnotesize
\tabstyle{1.00}{2}{1.2}{
\begin{tabular}{c|ccc|ccc|ccc}
\toprule
\multirow{2}{*}{\textbf{Dataset}}
& \multicolumn{3}{c|}{\textbf{1 Query}}
& \multicolumn{3}{c|}{\textbf{2 Queries}}
& \multicolumn{3}{c}{\textbf{4 Queries}} \\
\cmidrule(lr){2-4} \cmidrule(lr){5-7} \cmidrule(lr){8-10}
& \textbf{Balanced Acc} & \textbf{Kappa} & \textbf{Weighted F1}
& \textbf{Balanced Acc} & \textbf{Kappa} & \textbf{Weighted F1}
& \textbf{Balanced Acc} & \textbf{Kappa} & \textbf{Weighted F1} \\
\midrule
FACED
& 61.92 $\pm$ 0.47
& 57.00 $\pm$ 0.47
& 62.31 $\pm$ 0.47
& 62.31 $\pm$ 0.45
& 57.42 $\pm$ 0.50
& 62.65 $\pm$ 0.45
& \textbf{63.52} $\pm$ 0.37
& \textbf{58.83} $\pm$ 0.42
& \textbf{63.90} $\pm$ 0.43 \\

PhysioNet-MI
& 61.98 $\pm$ 0.53
& 49.30 $\pm$ 0.71
& 62.29 $\pm$ 0.57
& 63.02 $\pm$ 0.52
& 50.69 $\pm$ 0.69
& 63.27 $\pm$ 0.57
& \textbf{63.42} $\pm$ 0.90
& \textbf{51.22} $\pm$ 1.20
& \textbf{63.65} $\pm$ 0.96 \\

SEED-V
& \textbf{40.50} $\pm$ 0.37
& \textbf{25.78} $\pm$ 0.50
& \textbf{40.98} $\pm$ 0.47
& 40.18 $\pm$ 0.42
& 25.41 $\pm$ 0.44
& 40.69 $\pm$ 0.32
& 40.32 $\pm$ 0.47
& 25.48 $\pm$ 0.57
& 40.76 $\pm$ 0.44 \\
\midrule

\textit{Average}
& 54.80 $\pm$ 0.46
& 44.03 $\pm$ 0.56
& 55.19 $\pm$ 0.50
& 55.17 $\pm$ 0.46
& 44.51 $\pm$ 0.54
& 55.54 $\pm$ 0.45
& \textbf{55.75} $\pm$ 0.58
& \textbf{45.18} $\pm$ 0.73
& \textbf{56.10} $\pm$ 0.61 \\
\bottomrule
\end{tabular}
}
\end{table}

\paragraph{TemporalFormer Design.}

We further study the number of learnable query tokens in TemporalFormer, which produces the temporal-state representation used by Expert Selector for shared same-timestep routing. We vary the query number from $1$ to $2$ and $4$, with all other settings fixed.
As shown in Table~\ref{tab:temporalformer_query}, using more query tokens improves average downstream performance. The $4$-query setting achieves the best average Balanced Accuracy, Cohen's Kappa, and Weighted F1, outperforming the $1$-query setting by $+0.95\%$, $+1.15\%$, and $+0.91\$$, respectively. It performs best on FACED and PhysioNet-MI, while SEED-V slightly favors $1$ query, indicating task-dependent routing capacity. We therefore use $4$ queries as the default TemporalFormer setting in \textsc{Trace}.

\paragraph{Multi-scale Channel Positional Encoding.}
We next examine the necessity of the MS-ChPE introduced in Section~\ref{sec:method:encoder}. Removing channel PE leads to a consistent drop on both datasets, decreasing Balanced Accuracy by $3.72\%$ on FACED and $5.79\%$ on BCIC2020-3. This confirms that explicit channel-axis positional information is important for modeling multi-electrode EEG signals. Among single-scale designs, the wider kernel performs better than the narrow one. Compared with the multi-scale default, the $q_1=19$ setting is within $1.18\%$ on FACED and $1.58\%$ on BCIC2020-3, while the $q_1=5$ setting lags by $2.44\%$ and $2.99\%$, respectively. These results suggest that a wider receptive field over the channel axis helps capture montage-level spatial relations, and the multi-scale design further combines local and global channel patterns for more robust EEG embeddings.

\begin{table}[t]
\centering
\caption{Analysis of MS-ChPE with different settings.}
\label{tab:channel_pe_ablation}
\tabstyle{0.88}{6}{1.2}{
\begin{tabular}{l|cc|cc}
\toprule
\multirow{2}{*}{\textbf{MS-ChPE Design}}
& \multicolumn{2}{c|}{\textbf{FACED}}
& \multicolumn{2}{c}{\textbf{BCIC2020-3}} \\
\cmidrule(lr){2-3}\cmidrule(lr){4-5}
& \textbf{Balanced Acc} & \textbf{Kappa}
& \textbf{Balanced Acc} & \textbf{Kappa} \\
\midrule

Multi-scale $\{q_r\}_{r=1}^{R}=\{5,11,19\}$
& \textbf{0.6352} $\pm$ 0.0037
& \textbf{0.5883}
& \textbf{0.6315} $\pm$ 0.0055
& \textbf{0.5393} \\

Single-scale ($R{=}1,\ q_1{=}5$)
& 0.6108 $\pm$ 0.0078
& 0.5604
& 0.6016 $\pm$ 0.0122
& 0.5020 \\

Single-scale ($R{=}1,\ q_1{=}19$)
& 0.6234 $\pm$ 0.0066
& 0.5753
& 0.6157 $\pm$ 0.0118
& 0.5197 \\

\textit{w/o} Channel PE
& 0.5980 $\pm$ 0.0052
& 0.5455
& 0.5736 $\pm$ 0.0113
& 0.4670 \\

\bottomrule
\end{tabular}
}
\end{table}

\begin{table}[t]
\centering
\caption{Analysis of pre-training data composition.}
\label{tab:data_source_ablation_chbmit}
\tabstyle{0.98}{5}{1.2}{
\begin{tabular}{cccccc|ccc}
\toprule
\multicolumn{6}{c|}{\textbf{Pre-training Sources}}
& \multicolumn{3}{c}{\textbf{CHB-MIT}} \\
\cmidrule(lr){1-6}\cmidrule(lr){7-9}
\textbf{TUEG} & \textbf{HBN} & \textbf{Phys-MI} & \textbf{Phys-MA} & \textbf{SHU-MI} & \textbf{CHB-MIT}
& \textbf{Balanced Acc} & \textbf{AUC-PR} & \textbf{AUROC} \\
\midrule

\cmark & \xmark & \xmark & \xmark & \xmark & \xmark
& 58.91 $\pm$ 1.89
& 39.25 $\pm$ 10.81
& 89.46 $\pm$ 4.14 \\

\xmark & \cmark & \xmark & \xmark & \xmark & \xmark
& 70.35 $\pm$ 0.87
& 56.54 $\pm$ 8.69
& 90.72 $\pm$ 4.24 \\

\cmark & \xmark & \cmark & \cmark & \cmark & \cmark
& 65.88 $\pm$ 4.10
& 44.00 $\pm$ 6.43
& 89.85 $\pm$ 1.89 \\

\cmark & \cmark & \cmark & \cmark & \cmark & \cmark
& \textbf{74.53} $\pm$ 1.78
& \textbf{59.21} $\pm$ 7.95
& \textbf{92.20} $\pm$ 2.32 \\

\bottomrule
\end{tabular}
}
\end{table}

\paragraph{Effect of Multi-source Pre-training.}
We investigate the impact of pre-training data composition on downstream transfer to CHB-MIT.
As shown in Table~\ref{tab:data_source_ablation_chbmit}, adding task-related sources, including the CHB-MIT training split, improves over TUEG-only pre-training. Balanced Accuracy increases from $58.91\%$ to $65.88\%$, and AUC-PR increases from $39.25\%$ to $44.00\%$. This suggests that exposure to the target dataset's training split is beneficial for downstream CHB-MIT transfer.
The best performance is achieved by the full multi-source corpus, reaching $74.53\%$ Balanced Accuracy, $59.21\%$ AUC-PR, and $92.20\%$ AUROC. This indicates that target-dataset exposure alone is not sufficient. \textsc{Trace} benefits more from combining TUEG, HBN, and task-related EEG sources, suggesting that the gains of multi-source pre-training come from complementary EEG sources across clinical recordings, high-density healthy-population EEG, and task-related neural dynamics.
Additional SEED-V results in Appendix~\ref{sec:abl_data_source} show a consistent benefit from the full multi-source pre-training corpus.

\section{Conclusion and Limitations}
\label{sec:conclusion}

We presented \textsc{Trace}, an autoregressive EEG foundation model designed to learn transferable causal representations from heterogeneous multi-source EEG. By combining multi-horizon future-patch forecasting with Temporal Routing MoE blocks, \textsc{Trace} adapts computation across non-stationary temporal regimes while preserving the cross-channel coherence of EEG observations at each timestep. Across eight downstream benchmarks, \textsc{Trace} achieves leading results on several tasks and competitive results on the remaining motor imagery and clinical event classification benchmarks. Our ablations show that cross-channel temporal routing, the TemporalFormer design, expert scaling, and multi-source pre-training each contribute to downstream transfer. These results suggest that autoregressive pre-training can be effective for EEG representation learning when the prediction objective and architecture are aligned with the temporal and multi-channel structure of the signal.

\paragraph{Limitations.}
\label{sec:limitations}

Although \textsc{Trace} is evaluated across diverse public EEG datasets, its empirical scope is still limited by the available benchmarks, preprocessing choices, and dataset-specific label protocols. In particular, our downstream evaluation mainly focuses on classification-oriented BCI and clinical recognition tasks. We have not yet systematically validated \textsc{Trace} on regression-style prediction problems, such as forecasting continuous clinical variables or early-warning tasks that require detecting risk before an event occurs. These settings may require different evaluation protocols, prediction horizons, and calibration analyses, and we view them as important directions for future work. The pre-training corpus spans multiple channel counts, montages, and recording domains, but it may not cover all acquisition hardware, clinical populations, artifact patterns, or deployment conditions encountered in practice. As a result, performance on unseen institutions, rare neurological conditions, or substantially different electrode layouts should be validated before high-stakes use.

\textsc{Trace} also increases model and training complexity through its Temporal Routing MoE blocks. While sparse expert activation improves flexibility, the full expert pool adds parameters and pre-training cost, and the optimal number of experts or activated experts can vary across downstream tasks. Finally, this work focuses on representation quality and transfer performance rather than clinical interpretability, privacy-preserving training, or fairness across demographic groups. These aspects are important directions for future work, especially if EEG foundation models are used in real-world BCI or clinical monitoring systems.

{
\small
\bibliographystyle{plainnat}
\bibliography{reference}
}

\clearpage

\appendix

\section{Related Work}
\label{sec:related_work}

\paragraph{EEG Foundation Models.}
Inspired by the success of foundation models in vision and language, recent EEG research has moved from task-specific models toward general-purpose EEG foundation models. Existing efforts mainly follow two self-supervised pre-training paradigms. One line learns EEG representations through contrastive learning, where BENDR~\citep{kostas2021bendr} first demonstrated the effectiveness of contrastive pre-training on large-scale EEG recordings, and Brant-style models further extend representation learning across physiological signals~\citep{zhang2023brant,yuan2024brant2}. Another line focuses on reconstruction-based pre-training. BIOT~\citep{yang2023biot} tokenizes biosignals into unified sequences for cross-dataset pre-training; LaBraM~\citep{jiang2024large} learns a vector-quantized neural tokenizer for spectrum-based masked prediction; CBraMod~\citep{wang2025cbramod} decouples spatial and temporal dependencies with a criss-cross Transformer; and CodeBrain~\citep{ma2025codebrain} combines temporal and frequency tokenization to capture structured EEG patterns. More recent work has started to introduce autoregressive EEG pre-training: NeuroLM~\citep{jiang2024neurolm} feeds discrete EEG tokens into an LLM for multi-channel autoregression, while EEGPT~\citep{yue2024eegpt} studies next-signal prediction for an EEG generalist foundation model. \textsc{Trace} follows this emerging autoregressive direction, but focuses on causal next-patch forecasting with Cross-Channel Temporal Routing FFNs (CTR-FFNs) to model moment-to-moment non-stationarity while preserving cross-channel coherence.

\paragraph{Mixture of Experts.}
Mixture-of-Experts (MoE) increases model capacity without proportionally increasing computation by sparsely activating a subset of experts for each input~\citep{shazeer2017outrageously}. Efficient routing strategies, such as Top-$1$ routing in Switch Transformer~\citep{fedus2022switch} and Top-$2$ gating in GShard~\citep{lepikhin2020gshard}, have made MoE widely used in modern Transformer models. However, standard MoE architectures are mostly designed for language-like token sequences and route each token independently. This token-wise routing is not well aligned with EEG signals, where channels at the same timestep jointly describe an instantaneous brain state. Routing each patch independently may fragment this multi-channel coherence, while a uniform feed-forward block cannot adapt to the temporal non-stationarity of EEG dynamics. \textsc{Trace} introduces a CTR-FFN for autoregressive EEG pre-training: TemporalFormer summarizes the causal cross-channel temporal context, and an Expert Selector routes all channels at the same timestep to the same specialized expert subset. This enables expert specialization over evolving temporal contexts while preserving multi-channel EEG coherence.

\section{Dataset Description}
\label{sec:dataset_details}

We use a diverse collection of public EEG datasets for \textsc{Trace} pre-training and downstream evaluation. These datasets span large-scale clinical EEG, high-density healthy-population EEG, affective decoding, motor imagery, mental arithmetic, sleep staging, seizure detection, clinical event classification, and imagined speech. The collection covers heterogeneous recording domains, channel configurations, sampling rates, and temporal resolutions for learning and evaluating transferable EEG representations.

\paragraph{TUEG.}
The Temple University Hospital EEG Corpus (TUEG) is a large-scale public clinical EEG corpus collected in hospital settings~\citep{obeid2016temple}. It contains long-duration scalp EEG recordings from diverse clinical populations and covers routine, ambulatory, and continuous EEG studies. Following prior EEG foundation-model preprocessing protocols~\citep{jiang2024large,wang2025cbramod}, we use the $19$-channel international $10-20$ montage, apply band-pass filtering, notch filtering, resample the signals to $200$\,Hz, and segment recordings into $30$-second non-overlapping windows. TUEG provides large-scale clinical EEG background activity under real-world acquisition conditions.

\paragraph{HBN.}
The Healthy Brain Network (HBN) dataset provides high-density EEG recordings from children and adolescents across multiple experimental paradigms~\citep{alexander2017open}. Different from TUEG, which uses a $19$-channel clinical montage, HBN uses a $128$-channel HydroCel Geodesic Sensor Net and covers a healthy-population developmental cohort. This dataset introduces richer spatial coverage, different channel configurations, and non-clinical recording conditions. In our preprocessing pipeline, HBN recordings are resampled to $200$\,Hz and segmented into $30$-second windows while preserving the native $128$-channel layout.

\paragraph{Emotion Recognition.}
We conduct emotion recognition experiments on FACED and SEED-V. FACED is a fine-grained affective EEG dataset containing $32$-channel EEG recordings collected during affective video watching~\citep{chen2023faced}. It is formulated as a $9$-class emotion classification task covering multiple positive, negative, and neutral affective states. We follow the standard cross-subject evaluation protocol, resample the EEG signals to $200$\,Hz, and segment trials into $10$-second windows.

SEED-V is a $5$-class EEG emotion recognition dataset with happy, sad, neutral, disgust, and fear categories~\citep{liu2021comparing}. It contains $62$-channel multi-session EEG recordings from $16$ subjects, originally sampled at $1{,}000$\,Hz. Each subject participates in three sessions with $15$ trials per session. We split the $15$ trials in each session into training, validation, and test partitions of $5$ trials each, resample the signals to $200$\,Hz, and segment them into $1$-second windows.

\paragraph{Motor Imagery Classification.}
We evaluate motor imagery classification on PhysioNet-MI and SHU-MI. PhysioNet-MI contains $64$-channel EEG recordings from $109$ subjects performing motor execution and motor imagery tasks, originally sampled at $160$\,Hz~\citep{schalk2004bci2000}. We formulate it as a $4$-class classification task, resample the recordings to $200$\,Hz, and segment the signals into $4$-second windows. Following a subject-wise split, subjects $1-70$ are used for training, $71-89$ for validation, and $90-109$ for testing.

SHU-MI is a motor imagery dataset collected from $25$ subjects using a $32$-channel EEG cap, originally sampled at $250$\,Hz~\citep{ma2022large}. Each subject performs five sessions of left-hand and right-hand motor imagery. We formulate it as a binary classification task, resample the recordings to $200$\,Hz, and segment them into $4$-second non-overlapping windows. Following a subject-wise split, subjects $1-15$ are used for training, $16-20$ for validation, and $21-25$ for testing.

\paragraph{Mental Arithmetic.}
PhysioNet-MA contains EEG recordings from 36 subjects under two conditions: resting state and mental arithmetic~\citep{zyma2019electroencephalograms}. The original recordings consist of $72$ EDF files, corresponding to two recordings per subject, sampled at $500$\,Hz with $23$ electrodes. We retain $20$ standard $10-20$ channels, resample the signals to $200$\,Hz, and segment them into $5$-second non-overlapping windows. Following a subject-wise split, subjects $1-28$ are used for training, $29-32$ for validation, and $33-36$ for testing.

\paragraph{Sleep Staging.}
ISRUC is used for sleep stage classification~\citep{khalighi2016isruc}. It contains overnight polysomnographic recordings annotated according to standard sleep-stage categories: Wake, N1, N2, N3, and REM. Each EEG segment corresponds to a $30$-second epoch. We adopt subject-wise splits used in prior EEG foundation-model benchmarks: for ISRUC-S1, which contains $100$ subjects, we use $80$ subjects for training, $10$ for validation, and $10$ for testing; for ISRUC-S3, which contains $10$ subjects, we follow an $8:1:1$ subject-wise split.

\paragraph{Seizure Detection.}
CHB-MIT is a widely used pediatric scalp EEG dataset for seizure detection~\citep{shoeb2009application}. It contains long-term EEG recordings from subjects with epilepsy collected during clinical monitoring. We formulate the task as binary classification between seizure and non-seizure EEG segments. Following common preprocessing protocols, we use $16$ channels in the standard bipolar montage, resample recordings to $200$\,Hz, and segment them into $10$-second non-overlapping windows. We adopt a subject-wise split with chb01--chb20 for training, chb21-chb22 for validation, and chb23-chb24 for testing, with chb12, chb13, and chb17 excluded due to insufficient or noisy recordings.

\paragraph{EEG Event Classification.}
TUEV is used for clinical EEG event type classification~\citep{obeid2016temple}. The annotations cover six event categories: spike and sharp wave (SPSW), generalized periodic epileptiform discharges (GPED), periodic lateralized epileptiform discharges (PLED), eye movement (EYEM), artifact (ARTF), and background activity (BCKG). Following standard preprocessing, we select $16$ bipolar montage channels, resample the recordings to $200$\,Hz, and segment them into $5$-second windows. We follow the official train/test split and further divide the training subjects into an $8:2$ train/validation partition.

\paragraph{Imagined Speech Classification.}
BCI Competition 2020 Track 3 (BCIC2020-3) is used for imagined speech classification~\citep{jeong20222020}. The dataset contains $64$-channel EEG recordings collected while subjects silently imagine speaking five short words or phrases: ``hello'', ``help me'', ``stop'', ``thank you'', and ``yes''. It is formulated as a $5$-class classification task. We resample the recordings to $200$\,Hz and use the train, validation, and test splits provided by the competition organizers.

\section{Baselines and Metrics Description}
\label{app:baselines_metrics}

\subsection{Baselines}
\label{app:baselines}

Here, we introduce the details of the baselines used for performance evaluation.

\textbf{EEGNet}~\citep{lawhern2018eegnet} is a lightweight convolutional neural network designed for EEG decoding. It uses temporal convolution and depthwise spatial convolution to learn compact EEG representations.

\textbf{EEGConformer}~\citep{song2022eeg} is an EEG model that combines CNN-based local feature extraction with Transformer-based self-attention to model global temporal dependencies.

\textbf{CNN-Transformer}~\citep{peh2022eeg} is a hybrid architecture that uses CNN layers to extract local EEG features and a Transformer encoder to capture long-range temporal dependencies.

\textbf{FFCL}~\citep{li2022ffcl} is an EEG classification model that combines convolutional and recurrent modules, where CNNs capture local spatial-temporal patterns and recurrent layers model sequential dynamics.

\textbf{ST-Transformer}~\citep{song2021st} is a Transformer-based EEG model that learns spatial and temporal dependencies through attention mechanisms.

\textbf{BIOT}~\citep{yang2023biot} is a biosignal foundation model that tokenizes heterogeneous physiological signals into unified token sequences and learns transferable representations through cross-dataset pre-training.

\textbf{LaBraM}~\citep{jiang2024large} is a large EEG foundation model that learns neural tokens with a vector-quantized tokenizer and performs masked EEG token prediction for pre-training.

\textbf{CBraMod}~\citep{wang2025cbramod} is an EEG foundation model that uses criss-cross spatial-temporal attention and reconstruction-based pre-training to learn transferable EEG representations.

\textbf{CodeBrain}~\citep{ma2025codebrain} is an EEG foundation model that learns discrete EEG representations through temporal-frequency tokenization, aiming to capture structured EEG patterns for downstream transfer.

\subsection{Metrics}
\label{app:metrics}

In this section, we introduce the evaluation metrics adopted in this paper. Consistent with prior EEG foundation model evaluations~\citep{jiang2024large,wang2025cbramod,ma2025codebrain}, we report different metrics according to the task type.

\textbf{Balanced Accuracy} measures the average recall across all classes and is less affected by class imbalance than standard accuracy. It is reported for both binary and multi-class classification tasks.

\textbf{Cohen's Kappa} quantifies the agreement between predicted labels and ground-truth labels after correcting for chance agreement. This metric is particularly useful for multi-class EEG classification benchmarks with imbalanced class distributions.

\textbf{Weighted F1} computes the class-wise F1 score and averages it according to class support. It provides a precision-recall summary for multi-class tasks where different classes may have different numbers of samples.

\textbf{AUC-PR} denotes the area under the precision-recall curve. It is especially informative for imbalanced binary detection tasks, where the positive class is rare.

\textbf{AUROC} denotes the area under the receiver operating characteristic curve and evaluates threshold-independent discrimination between positive and negative classes.

\section{More Details for Experimental Settings on Pre-Training}
\label{app:pretrain_details}

We next provide additional details on the pre-training corpus, model configuration, and optimization setup used for \textsc{Trace} pre-training.

\paragraph{Pre-training corpus.}
We construct the pre-training corpus from six EEG sources: two large-scale sources, TUEG and HBN, together with four task-related datasets. TUEG provides large-scale 19-channel clinical EEG, while HBN provides $128$-channel high-density EEG from a healthy-population cohort. The task-related sources further cover seizure monitoring, motor imagery, and mental arithmetic recordings. For datasets that also appear in downstream evaluation, only the training-split subjects are included in the pre-training corpus; validation and test subjects are strictly held out from pre-training.
The sources and preprocessing details are summarized in Table~\ref{tab:pretrain_sources}. The sample counts in the table indicate the total number of available preprocessed segments for each source; for downstream-overlapping datasets, pre-training uses only the training split.

\begin{table}[h]
\centering
\caption{{Pre-training corpus and preprocessing details.}}
\label{tab:pretrain_sources}
\tabstyle{0.85}{6}{1.2}{
\begin{tabular}{llccc}
\toprule
\textbf{Dataset} & \textbf{EEG Domain} & \textbf{\# Channels} & \textbf{\# Samples} & \textbf{Window} \\
\midrule
TUEG & Clinical EEG & 19 & 1.1M & 30s \\
HBN & Healthy-population EEG & 128 & 132K & 30s \\
\midrule
CHB-MIT & Seizure detection & 16 & 326,993 & 10s \\
PhysioNet-MI & Motor imagery & 64 & 9,837 & 4s \\
SHU-MI & Motor imagery & 32 & 11,988 & 4s \\
PhysioNet-MA & Stress detection & 20 & 1,343 & 5s \\
\bottomrule
\end{tabular}
}
\end{table}

\paragraph{Model configuration.}
The architecture hyperparameters of \textsc{Trace} are summarized in Table~\ref{tab:model_config}. The main configuration uses $N{=}64$ total experts and activates $K{=}8$ experts at each temporal step.

 \begin{table}[h]
  \caption{Hyperparameters of \textsc{Trace} architecture.}
  \label{tab:model_config}
  \centering
\tabstyle{0.8}{7}{1.2}{
  \begin{tabular}{ll}
    \toprule
    \textbf{Hyperparameters}                          & \textbf{Values} \\
    \midrule
    \textbf{Patch embedding}                          & \\
    \quad Temporal kernel set $|\mathcal{Q}_{\mathrm{temp}}|$ & $3$ \\
    \quad Temporal kernels $\mathcal{Q}_{\mathrm{temp}}$      & $\{25,\,49,\,99\}$ \\
    \quad CNN stride                                  & $25$ \\
    \quad Filters per branch                          & $8$ \\
    \quad Spectral branch                             & FFT log-amplitude \\
    \quad Spectral frequency bins                     & $101$ \\
    \quad Fusion                                      & sigmoid-gated \\
    \quad Patch embedding dimension $d$               & $200$ \\
    \midrule
    \textbf{Multi-Scale Channel Positional Encoding (MS-ChPE)} & \\
    \quad Channel-kernel branches $R$                 & $3$ depth-wise Conv2d \\
    \quad Channel kernels $\{q_r\}_{r=1}^{R}$         & $\{5,\,11,\,19\}$ \\
    \quad Time kernel                                 & $1$ \\
    \midrule
    \textbf{Transformer backbone}                     & \\
    \quad Layers $L$                                  & $12$ \\
    \quad Hidden size $d$                             & $200$ \\
    \quad FFN intermediate size                       & $800$ \\
    \quad Attention heads                             & $8$ \\
    \quad KV heads                                    & $8$ \\
    \quad Positional encoding                         & RoPE ($\theta{=}10{,}000$) \\
    \quad Activation                                  & SiLU \\
    \quad Normalization                               & RMSNorm ($\epsilon{=}1\mathrm{e}{-}6$) \\
    \midrule
    \textbf{Cross-Channel Temporal Routing FFN (CTR-FFN)} & \\
    \quad Specialized experts $N$                     & $64$ \\
    \quad Activated specialized experts $K$           & $8$ \\
    \quad Shared expert $\operatorname{Expert}_{N+1}$ & $1$ \\
    \quad Routing context                             & TemporalFormer \\
    \quad Expert selection                            & Top-$K$ Expert Selector \\
    \quad TemporalFormer query tokens $m$             & $4$ \\
    \quad TemporalFormer attention heads              & $4$ \\
    \bottomrule
  \end{tabular}
  }
\end{table}

\paragraph{Optimization setup.}
The pre-training optimization hyperparameters are summarized in Table~\ref{tab:pretrain_hparams}.

\begin{table}[h]
  \caption{Hyperparameters of pre-training.}
  \label{tab:pretrain_hparams}
  \centering
\tabstyle{0.55}{8}{1.2}{
  \begin{tabular}{ll}
    \toprule
    \textbf{Hyperparameters}                              & \textbf{Values} \\
    \midrule
    \textbf{Schedule}                                     & \\
    \quad Epochs                                          & $30$ \\
    \quad Batch size                                      & $256$ \\
    \quad Mixed precision                                 & bf16 \\
    \midrule
    \textbf{Optimizer}                                    & \\
    \quad Linear params (dim\,$\geq$\,$2$)                & Muon, $\mathrm{lr}{=}1\mathrm{e}{-}3$ \\
    \quad Embeddings / LN / biases                        & AdamW, $\mathrm{lr}{=}2\mathrm{e}{-}5$ \\
    \quad Weight decay                                    & $1\mathrm{e}{-}2$ \\
    \quad LR schedule                                     & cosine decay \\
    \quad Warmup steps                                    & $2{,}000$ \\
    \quad Gradient clipping (max norm)                    & $1.0$ \\
    \midrule
    \textbf{Forecasting objective}                        & \\
    \quad Horizons $\mathcal{H}$                          & $\{1,\,2,\,4\}$ \\
    \quad Huber threshold $\delta$                        & $1.0$ \\
    \quad Autoregressive loss $\mathcal{L}_{\mathrm{AR}}$ & unit weight \\
    \midrule
    \textbf{Auxiliary losses}                             & \\

    \quad Expert-balancing weight $\lambda_{\mathrm{aux}}$ & $1\mathrm{e}{-}2$ \\
    \bottomrule
  \end{tabular}
  }
\end{table}

\section{Fine-tuning Settings on Downstream Tasks}
\label{app:downstream_details}

\paragraph{Downstream Tasks.}
To evaluate the transferability of \textsc{Trace}, we fine-tune the pretrained model on eight downstream EEG datasets covering emotion recognition, motor imagery classification, sleep staging, seizure detection, imagined speech classification, and EEG event classification. For all downstream datasets, EEG signals are resampled to 200 Hz, and each EEG patch is set to 1s, corresponding to 200 data points, following the same tokenization setting used in pre-training. Details of the downstream tasks and datasets are summarized in Table~\ref{tab:downstream_overview}.

\begin{table}[H]
\centering
\caption{Overview of downstream BCI tasks and datasets.}
\label{tab:downstream_overview}
\tabstyle{0.92}{6}{1.2}{
\begin{tabular}{llcccc}
\toprule
\textbf{Downstream Tasks} & \textbf{Dataset} & \textbf{\# Channels} & \textbf{\# Samples} & \textbf{Window} & \textbf{Label} \\
\midrule
\multirow[t]{2}{*}{Emotion Recognition}
& FACED & 32 & 10,332 & 10s & 9-class \\
& SEED-V & 62 & 117,744 & 1s & 5-class \\

\midrule
\multirow[t]{2}{*}{Motor Imagery Classification}
& PhysioNet-MI & 64 & 9,837 & 4s & 4-class \\
& SHU-MI & 32 & 11,988 & 4s & 2-class \\

\midrule
Sleep Staging
& ISRUC & 6 & 89,240 & 30s & 5-class \\

Seizure Detection
& CHB-MIT & 16 & 326,993 & 10s & 2-class \\

Imagined Speech Classification
& BCIC2020-3 & 64 & 6,000 & 3s & 5-class \\

Event Type Classification
& TUEV & 16 & 112,491 & 5s & 6-class \\
\bottomrule
\end{tabular}
}
\end{table}

\paragraph{Fine-tuning setup.}
For each downstream task, we attach a lightweight three-layer MLP classification head on top of the pretrained \textsc{Trace} encoder and fine-tune the full model end-to-end. The fine-tuning hyperparameters are selected for each dataset based on the task type and dataset scale. Table~\ref{tab:finetune_hparams} lists the learning rate, weight decay, dropout rate, and batch size used for each downstream benchmark.

\begin{table}[h]
  \caption{Fine-tuning hyperparameters for downstream tasks.}
  \label{tab:finetune_hparams}
  \centering
  \tabstyle{0.75}{7}{1.2}{
  \begin{tabular}{lcccc}
    \toprule
    \textbf{Dataset} & \textbf{Learning rate} & \textbf{Weight decay} & \textbf{Dropout} & \textbf{Batch size} \\
    \midrule
    FACED        & $2\mathrm{e}{-}3$ & $5\mathrm{e}{-}2$ & $0.05$ & $16$  \\
    SEED-V       & $2\mathrm{e}{-}3$ & $5\mathrm{e}{-}2$ & $0.10$ & $256$ \\
    PhysioNet-MI & $5\mathrm{e}{-}3$ & $5\mathrm{e}{-}2$ & $0.10$ & $64$  \\
    SHU-MI       & $1\mathrm{e}{-}3$ & $5\mathrm{e}{-}3$ & $0.10$ & $32$  \\
    ISRUC        & $5\mathrm{e}{-}3$ & $5\mathrm{e}{-}2$ & $0.10$ & $16$  \\
    CHB-MIT      & $2\mathrm{e}{-}3$ & $5\mathrm{e}{-}3$ & $0.10$ & $64$  \\
    BCIC2020-3   & $5\mathrm{e}{-}3$ & $5\mathrm{e}{-}2$ & $0.10$ & $32$  \\
    TUEV         & $2\mathrm{e}{-}3$ & $5\mathrm{e}{-}2$ & $0.10$ & $32$  \\
    \bottomrule
  \end{tabular}
}
\end{table}

\section{More Results on Other Downstream BCI Tasks}
\label{app:more_downstream_results}

We provide additional downstream results on SHU-MI and TUEV to further evaluate the transfer ability of \textsc{Trace} across motor imagery and EEG event classification tasks.

\paragraph{SHU-MI.}
As shown in Table~\ref{tab:shumi_more_results}, CodeBrain achieves the best overall performance on SHU-MI, reaching $0.6431$ Balanced Accuracy, $0.7166$ AUC-PR, and $0.7124$ AUROC. \textsc{Trace} remains competitive among EEG foundation models, obtaining $0.6329$ Balanced Accuracy, $0.6978$ AUC-PR, and $0.6988$ AUROC. Notably, its AUROC matches CBraMod ($0.6988$), while its Balanced Accuracy remains higher than BIOT and LaBraM-Base. Compared with task-specific baselines such as EEGConformer and ST-Transformer, \textsc{Trace} also shows a clear improvement in Balanced Accuracy and AUC-PR. These results suggest that although \textsc{Trace} does not achieve the best result on SHU-MI, it maintains strong transfer performance on motor imagery classification.

\begin{table}[H]
\centering
\caption{Performance comparison on the SHU-MI (2-Class) dataset.}
\label{tab:shumi_more_results}
\tabstyle{0.78}{7}{1.2}{

\begin{tabular}{l|ccc}
\toprule
\textbf{Method} & \textbf{Balanced Accuracy} & \textbf{AUC-PR} & \textbf{AUROC} \\
\midrule
EEGNet
& 0.5889 $\pm$ 0.0177 & 0.6311 $\pm$ 0.0142 & 0.6283 $\pm$ 0.0152 \\
EEGConformer
& 0.5900 $\pm$ 0.0107 & 0.6370 $\pm$ 0.0093 & 0.6351 $\pm$ 0.0101 \\
CNN-Transformer
& 0.5975 $\pm$ 0.0169 & 0.6412 $\pm$ 0.0076 & 0.6323 $\pm$ 0.0082 \\
FFCL
& 0.5692 $\pm$ 0.0252 & 0.5943 $\pm$ 0.0172 & 0.6014 $\pm$ 0.0168 \\
ST-Transformer
& 0.5992 $\pm$ 0.0206 & 0.6394 $\pm$ 0.0122 & 0.6431 $\pm$ 0.0111 \\
\midrule
BIOT
& 0.6179 $\pm$ 0.0183 & 0.6770 $\pm$ 0.0119 & 0.6609 $\pm$ 0.0127 \\
LaBraM-Base
& 0.6166 $\pm$ 0.0192 & 0.6761 $\pm$ 0.0083 & 0.6604 $\pm$ 0.0091 \\
CBraMod
& \secondbest{0.6370} $\pm$ 0.0151
& \secondbest{0.7139} $\pm$ 0.0088
& \secondbest{0.6988} $\pm$ 0.0068 \\
CodeBrain
& \best{0.6431} $\pm$ 0.0066
& \best{0.7166} $\pm$ 0.0106
& \best{0.7124} $\pm$ 0.0050 \\
\midrule
\textbf{\textsc{Trace} (Ours)}
& 0.6329 $\pm$ 0.0061
& 0.6978 $\pm$ 0.0059
& \secondbest{0.6988} $\pm$ 0.0048 \\
\bottomrule
\end{tabular}
}
\end{table}

\paragraph{TUEV.}
As shown in Table~\ref{tab:tuev_more_results}, TUEV is a challenging six-class clinical EEG event classification benchmark. \textsc{Trace} obtains $0.6340$ Balanced Accuracy, $0.6358$ Cohen's Kappa, and $0.8113$ Weighted F1. Although the strongest results on this dataset are achieved by CBraMod and CodeBrain, \textsc{Trace} substantially outperforms task-specific baselines such as EEGConformer and ST-Transformer, and also improves over BIOT across all three metrics. Compared with BIOT, \textsc{Trace} increases Balanced Accuracy by $+10.59\%$, Cohen's Kappa by $+10.85\%$, and Weighted F1 by $+6.21\%$ percentage points. These results indicate that \textsc{Trace} transfers effectively to clinical event classification, while also showing that TUEV remains a demanding benchmark where reconstruction- or tokenization-based EEG foundation models can retain an advantage.

\begin{table}[H]
\centering
\caption{Performance comparison on the TUEV (6-Class) dataset.}
\label{tab:tuev_more_results}
\tabstyle{0.80}{7}{1.2}{
\begin{tabular}{l|ccc}
\toprule
\textbf{Method} & \textbf{Balanced Accuracy} & \textbf{Cohen's Kappa} & \textbf{Weighted F1} \\
\midrule
EEGNet
& 0.3876 $\pm$ 0.0143 & 0.3577 $\pm$ 0.0155 & 0.6539 $\pm$ 0.0120 \\
EEGConformer
& 0.4074 $\pm$ 0.0164 & 0.3967 $\pm$ 0.0195 & 0.6983 $\pm$ 0.0152 \\
CNN-Transformer
& 0.4087 $\pm$ 0.0161 & 0.3815 $\pm$ 0.0134 & 0.6854 $\pm$ 0.0293 \\
FFCL
& 0.3979 $\pm$ 0.0104 & 0.3732 $\pm$ 0.0188 & 0.6783 $\pm$ 0.0120 \\
ST-Transformer
& 0.3984 $\pm$ 0.0228 & 0.3765 $\pm$ 0.0306 & 0.6823 $\pm$ 0.0190 \\
\midrule
BIOT
& 0.5281 $\pm$ 0.0225 & 0.5273 $\pm$ 0.0249 & 0.7492 $\pm$ 0.0082 \\
LaBraM-Base
& 0.6409 $\pm$ 0.0065 & 0.6637 $\pm$ 0.0093 & 0.8312 $\pm$ 0.0052 \\
CBraMod
& \textbf{0.6671} $\pm$ 0.0107
& 0.6772 $\pm$ 0.0096
& 0.8342 $\pm$ 0.0064 \\
CodeBrain
& 0.6428 $\pm$ 0.0062
& \textbf{0.6912} $\pm$ 0.0101
& \textbf{0.8362} $\pm$ 0.0048 \\
\midrule
\textbf{\textsc{Trace} (Ours)}
& 0.6340 $\pm$ 0.0036
& 0.6358 $\pm$ 0.0244
& 0.8113 $\pm$ 0.0114 \\
\bottomrule
\end{tabular}
}
\end{table}

\section{Implementation Details}
\label{app:implementation_details}
\label{sec:impl_details}

\subsection{Causal Spatial-Temporal Attention}
\label{app:csta}

This section provides the detailed implementation of the causal spatial--temporal attention (CSTA) module used in each TR-MoE block. Let the input to the $l$-th block be
$\mathbf{H}^{l}\in\mathbb{R}^{C\times n\times d}$, where $C$ is the number of EEG channels, $n$ is the number of temporal patches, and $d$ is the hidden dimension. CSTA factorizes full attention over the $C n$ patch tokens into two structured attention branches: a spatial branch over channels at the same temporal step and a causal temporal branch over past patches within each channel. This design captures both instantaneous cross-channel coupling and within-channel temporal dynamics while preserving the autoregressive constraint.

\paragraph{Spatial attention branch.}
For each temporal step $j$, we collect the channel tokens
$\mathbf{H}^{l}_{:,j}\in\mathbb{R}^{C\times d}$ and apply multi-head self-attention along the channel axis:
\begin{equation}
    \mathbf{S}^{l}_{:,j}
    =
    \operatorname{MHA}_{\mathrm{sp}}
    \left(
        \mathbf{H}^{l}_{:,j},
        \mathbf{H}^{l}_{:,j},
        \mathbf{H}^{l}_{:,j}
    \right).
\end{equation}

The attention weights are computed only among channels that belong to the same temporal patch index $j$. Therefore, the spatial branch models global electrode interactions at each timestep without using information from any future temporal patch. In practice, this branch is implemented by reshaping the tensor from $(B,C,n,d)$ to $(B n,C,d)$ and applying standard multi-head attention with sequence length $C$.

\paragraph{Causal temporal attention branch.}
For each channel $i$, we collect its temporal sequence
$\mathbf{H}^{l}_{i,:}\in\mathbb{R}^{n\times d}$ and apply multi-head self-attention along the patch axis with a lower-triangular causal mask:
\begin{equation}
    \mathbf{T}^{l}_{i,:}
    =
    \operatorname{MHA}_{\mathrm{time}}
    \left(
        \mathbf{H}^{l}_{i,:},
        \mathbf{H}^{l}_{i,:},
        \mathbf{H}^{l}_{i,:};
        \mathbf{M}_{\mathrm{causal}}
    \right),
\end{equation}
where $\mathbf{M}_{\mathrm{causal}}(j,j')=0$ if $j'\leq j$ and $-\infty$ otherwise. Thus, the output token at temporal step $j$ can attend only to patches $\{1,\ldots,j\}$ from the same channel. This branch is implemented by reshaping $(B,C,n,d)$ to $(B C,n,d)$ and applying standard masked multi-head attention with sequence length $n$.

\paragraph{Branch fusion.}
The two attention outputs are reshaped back to $(B,C,n,d)$ and fused with an output projection:
\begin{equation}
    \operatorname{CSTA}(\mathbf{H}^{l})_{i,j}
    =
    \mathbf{W}_{\mathrm{o}}
    \left[
        \mathbf{S}^{l}_{i,j};
        \mathbf{T}^{l}_{i,j}
    \right],
\end{equation}
where $[\cdot;\cdot]$ denotes feature concatenation and $\mathbf{W}_{\mathrm{o}}\in\mathbb{R}^{2d\times d}$ projects the fused representation back to the model dimension. The projected output is then used in the pre-norm residual update in the TR-MoE block, as defined in Section~\ref{sec:method:trmoe}.

CSTA is causal because the spatial branch only mixes channels at the same temporal step and the temporal branch uses a causal mask along the patch axis. Compared with full attention over all $Cn$ tokens, which has complexity $O(C^{2}n^{2})$, the factorized implementation has complexity $O(nC^{2}+Cn^{2})$. This is important for EEG pre-training because both the channel count and sequence length can vary substantially across datasets.

\section{More Experimental Results}
\label{sec:more_experiments}

\subsection{Effect of Multi-Horizon Forecasting}
\label{app:horizon}

We first examine the design of the autoregressive forecasting objective. Instead of predicting only the next EEG patch, \textsc{Trace} can be trained to predict future patches at multiple temporal offsets. We compare three horizon sets, $\mathcal{H}=\{1\}$, $\mathcal{H}=\{1,2\}$, and $\mathcal{H}=\{1,2,4\}$.

\begin{table}[H]
\centering
\caption{Anlysis of the multi-horizon forecasting objective.
The best result in each dataset block is highlighted in \textbf{bold}.}
\label{tab:horizon_ablation}
\tabstyle{0.85}{7}{1.2}{

\begin{tabular}{c|l|ccc}
\toprule
\textbf{Dataset} & \textbf{Horizon Set} & \textbf{Balanced Acc} & \textbf{Cohen's Kappa} & \textbf{Weighted F1} \\
\midrule

\multirow{3}{*}{\centering\textbf{FACED}}
& $\mathcal{H}=\{1\}$
& 62.14 $\pm$ 0.35 & 57.25 $\pm$ 0.33 & 62.47 $\pm$ 0.36 \\
& $\mathcal{H}=\{1,2\}$
& 61.31 $\pm$ 1.39 & 56.25 $\pm$ 1.63 & 61.57 $\pm$ 1.44 \\
& $\mathcal{H}=\{1,2,4\}$
& \textbf{62.83} $\pm$ 0.59
& \textbf{58.07} $\pm$ 0.69
& \textbf{63.17} $\pm$ 0.58 \\

\midrule

\multirow{3}{*}{\centering\textbf{SEED-V}}
& $\mathcal{H}=\{1\}$
& \textbf{40.95} $\pm$ 0.46
& \textbf{26.51} $\pm$ 0.76
& \textbf{41.64} $\pm$ 0.65 \\
& $\mathcal{H}=\{1,2\}$
& 40.73 $\pm$ 0.51 & 26.16 $\pm$ 0.71 & 41.37 $\pm$ 0.55 \\
& $\mathcal{H}=\{1,2,4\}$
& 40.63 $\pm$ 0.33 & 25.94 $\pm$ 0.42 & 41.19 $\pm$ 0.35 \\

\midrule

\multirow{3}{*}{\centering\textbf{PhysioNet-MI}}
& $\mathcal{H}=\{1\}$
& 63.07 $\pm$ 1.45 & 50.75 $\pm$ 1.93 & 63.28 $\pm$ 1.41 \\
& $\mathcal{H}=\{1,2\}$
& 63.49 $\pm$ 0.67 & 51.31 $\pm$ 0.89 & 63.71 $\pm$ 0.76 \\
& $\mathcal{H}=\{1,2,4\}$
& \textbf{63.62} $\pm$ 0.30
& \textbf{51.49} $\pm$ 0.40
& \textbf{63.90} $\pm$ 0.25 \\

\midrule

\multirow{3}{*}{\centering\textbf{Average}}
& $\mathcal{H}=\{1\}$
& 55.39 $\pm$ 0.75 & 44.84 $\pm$ 1.01 & 55.80 $\pm$ 0.81 \\
& $\mathcal{H}=\{1,2\}$
& 55.18 $\pm$ 0.86 & 44.57 $\pm$ 1.08 & 55.55 $\pm$ 0.92 \\
& $\mathcal{H}=\{1,2,4\}$
& \textbf{55.69} $\pm$ 0.41
& \textbf{45.17} $\pm$ 0.50
& \textbf{56.09} $\pm$ 0.39 \\

\bottomrule
\end{tabular}
}
\end{table}

As shown in Table~\ref{tab:horizon_ablation}, using $\mathcal{H}=\{1,2,4\}$ gives the best average performance across the three downstream datasets. Compared with the single-step objective $\mathcal{H}=\{1\}$, the multi-horizon objective improves the average Balanced Accuracy from $55.39\%$ to $55.69\%$, Cohen's Kappa from $44.84\%$ to $45.17\%$, and Weighted F1 from $55.80\%$ to $56.09\%$. The gain is most evident on FACED and PhysioNet-MI. On FACED, $\mathcal{H}=\{1,2,4\}$ improves Balanced Accuracy by $+0.69$ percentage points and Weighted F1 by $+0.70$ percentage points over $\mathcal{H}=\{1\}$. On PhysioNet-MI, it also achieves the best result across all metrics, with a smaller standard deviation than the single-step setting. SEED-V shows a slightly different pattern, where the single-step horizon performs best. This suggests that the optimal temporal prediction scale can vary across downstream tasks, while combining multiple horizons provides the strongest average transfer.

\begin{figure}[H]
  \centering
  \includegraphics[width=0.55\linewidth]{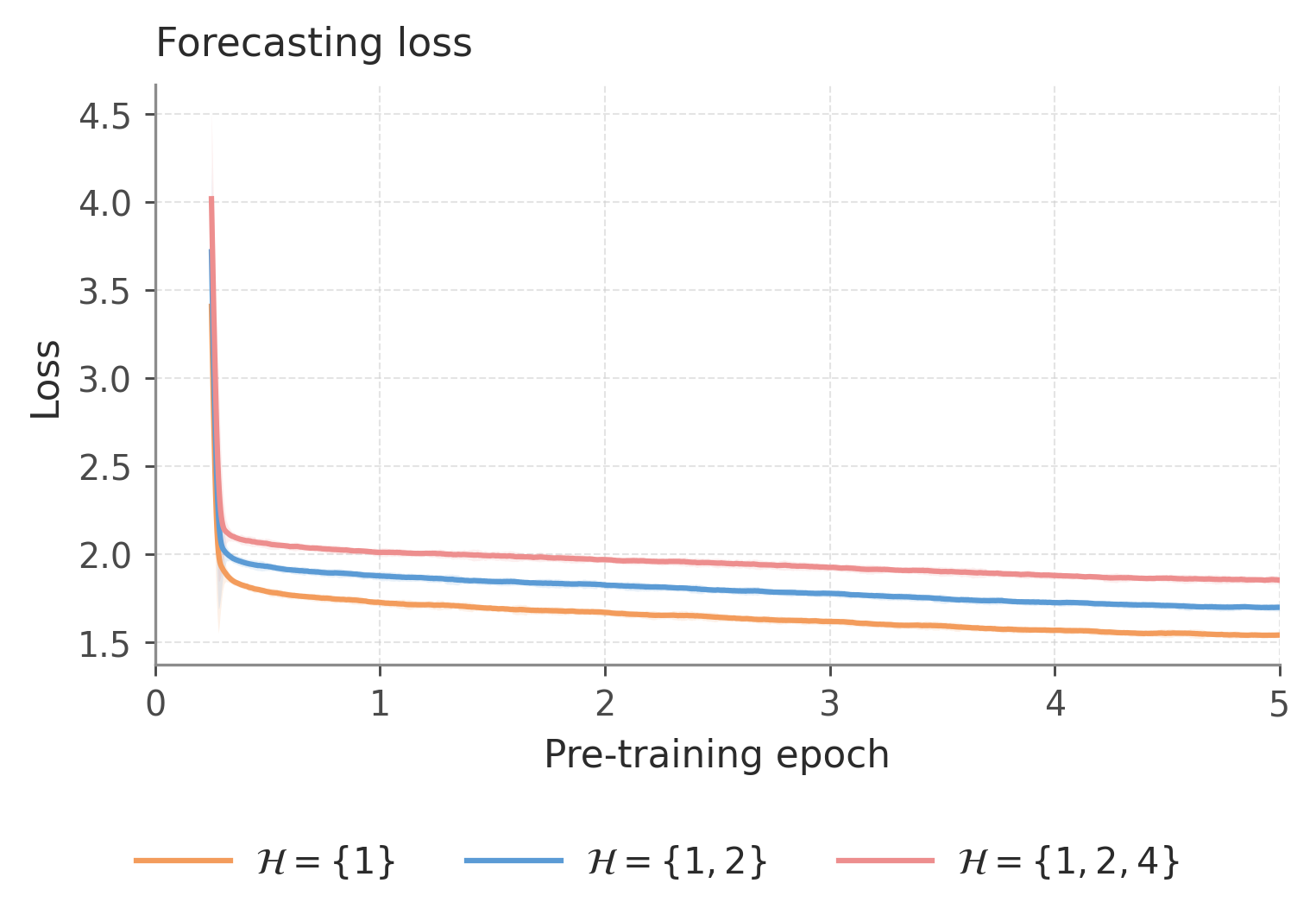}
  \caption{{Pre-training loss dynamics} under different horizon sets.}
  \label{fig:horizon_loss}
\end{figure}

Figure~\ref{fig:horizon_loss} further compares the pre-training dynamics under different horizon sets. Adding longer horizons increases the forecasting loss, which is expected because farther future patches are harder to predict from causal context. Importantly, this higher forecasting loss does not indicate weaker downstream transfer: $\mathcal{H}=\{1,2,4\}$ shows the largest forecasting loss but achieves the best average performance in Table~\ref{tab:horizon_ablation}. This suggests that multi-horizon forecasting creates a more challenging but more informative temporal prediction task.

\subsection{Effect of Multi-source Pre-training}
\label{sec:abl_data_source}

We evaluate the impact of pre-training data composition by comparing four source combinations, ranging from single-source pre-training to the full multi-source corpus. This experiment aims to determine whether source diversity, channel configuration, and task coverage improve downstream transfer.

As shown in Table~\ref{tab:data_source_ablation_seedv}, the full multi-source setting achieves the best performance on SEED-V across B-Acc, Kappa, and F1-W. This highlights two key findings. First, pre-training source domain matters. HBN-only slightly outperforms TUEG-only despite using fewer samples, suggesting that high-density 128-channel healthy-population EEG provides complementary information to the 19-channel clinical EEG in TUEG. Second, task-oriented sources are most effective when combined with large-scale EEG corpora. The 4ds-only setting performs the worst, while the full setting achieves the strongest result after adding the task-oriented datasets to TUEG and HBN. These results suggest that \textsc{Trace} benefits from pre-training on EEG data that spans different channel counts, acquisition conditions, and task domains, rather than relying on a single source or merely increasing sample count.

\begin{table}[H]
\centering
\caption{Analysis of pre-training data composition for SEED-V.}
\label{tab:data_source_ablation_seedv}

\tabstyle{1.00}{5}{1.2}{
\begin{tabular}{cccccc|ccc}
\toprule
\multicolumn{6}{c|}{\textbf{Pre-training Sources}}
& \multicolumn{3}{c}{\textbf{Downstream on SEED-V}} \\
\cmidrule(lr){1-6}\cmidrule(lr){7-9}
\textbf{TUEG} & \textbf{HBN} & \textbf{Phys-MI} & \textbf{Phys-MA} & \textbf{SHU-MI} & \textbf{CHB-MIT}
& \textbf{Balanced Acc} & \textbf{Cohen's Kappa} & \textbf{Weighted F1} \\
\midrule

\cmark & \xmark & \xmark & \xmark & \xmark & \xmark
& 41.10 $\pm$ 0.08
& 26.57 $\pm$ 0.14
& 41.69 $\pm$ 0.15 \\

\xmark & \cmark & \xmark & \xmark & \xmark & \xmark
& 41.24 $\pm$ 0.55
& 26.84 $\pm$ 0.73
& 42.06 $\pm$ 0.60 \\

\xmark & \xmark & \cmark & \cmark & \cmark & \cmark
& 40.81 $\pm$ 0.61
& 26.33 $\pm$ 0.52
& 41.42 $\pm$ 0.51 \\

\cmark & \cmark & \cmark & \cmark & \cmark & \cmark
& \textbf{41.86} $\pm$ 0.32
& \textbf{27.59} $\pm$ 0.30
& \textbf{42.53} $\pm$ 0.24 \\

\bottomrule
\end{tabular}
}
\end{table}

As shown in Table~\ref{tab:data_source_ablation_seedv}, the choice of pre-training data has a clear impact on downstream transfer. The mixed-source corpus provides the strongest overall transfer, suggesting that diverse EEG sources help \textsc{Trace} learn more general temporal representations. We also observe task-dependent transfer patterns, where sources closer to the target task tend to provide stronger gains on the corresponding downstream benchmark.

\subsection{Effect of CTR-FFN}
\label{sec:abl_routing}

We investigate the essential design of \textsc{Trace}, \ie, the CTR-FFN. We evaluate the necessity of this core architectural design by comparing the full \textsc{Trace} model against three variants: 1) a Non-MoE baseline where the CTR-FFN is replaced with a standard FFN, 2) a Vanilla MoE version where each patch selects its own experts independently via a token-wise router, and 3) a Mean Temporal variant where the routing context at each timestep is obtained by mean-pooling channel embeddings instead of using TemporalFormer. All experiments are conducted on three datasets: FACED, PhysioNet-MI, and SEED-V. As shown in Table~\ref{tab:moe_routing}, our proposed CTR-FFN consistently outperforms all baseline variants across the evaluated datasets.
\begin{table}[H]
\centering
\caption{Effect of CTR-FFN.}
\label{tab:moe_routing}
\tabstyle{0.90}{7}{1.2}{

\begin{tabular}{c|l|ccc}
\toprule
\textbf{Dataset} & \textbf{Routing Design} & \textbf{Balanced Accuracy} & \textbf{Cohen's Kappa} & \textbf{Weighted F1} \\
\midrule

\multirow{4}{*}{\shortstack[c]{\textbf{FACED}\\\textbf{9-Class}}}
& Non-MoE
& 62.44 $\pm$ 0.68 & 57.63 $\pm$ 1.16 & 62.91 $\pm$ 1.06 \\
& Vanilla MoE
& 62.39 $\pm$ 1.04 & 57.55 $\pm$ 0.79 & 62.66 $\pm$ 0.65 \\
& Mean Temporal
& 63.45 $\pm$ 0.53 & 58.76 $\pm$ 0.61 & 63.75 $\pm$ 0.59 \\
& \textbf{\textsc{Trace} (Ours)}
& \textbf{63.52} $\pm$ 0.37
& \textbf{58.83} $\pm$ 0.42
& \textbf{63.90} $\pm$ 0.43 \\

\midrule

\multirow{4}{*}{\shortstack[c]{\textbf{PhysioNet-MI}\\\textbf{4-Class}}}
& Non-MoE
& 62.28 $\pm$ 1.45 & 49.70 $\pm$ 1.93 & 62.53 $\pm$ 1.43 \\
& Vanilla MoE
& 62.76 $\pm$ 0.70 & 50.34 $\pm$ 0.93 & 63.06 $\pm$ 0.65 \\
& Mean Temporal
& 63.00 $\pm$ 0.84 & 50.66 $\pm$ 1.13 & 63.30 $\pm$ 0.87 \\
& \textbf{\textsc{Trace} (Ours)}
& \textbf{63.42} $\pm$ 0.90
& \textbf{51.22} $\pm$ 1.20
& \textbf{63.65} $\pm$ 0.96 \\

\midrule

\multirow{4}{*}{\shortstack[c]{\textbf{SEED-V}\\\textbf{5-Class}}}
& Non-MoE
& 39.82 $\pm$ 0.65 & 24.86 $\pm$ 0.96 & 40.27 $\pm$ 0.79 \\
& Vanilla MoE
& 38.43 $\pm$ 0.87 & 22.91 $\pm$ 1.26 & 38.75 $\pm$ 1.03 \\
& Mean Temporal
& 39.42 $\pm$ 0.50 & 24.18 $\pm$ 0.59 & 39.82 $\pm$ 0.46 \\
& \textbf{\textsc{Trace} (Ours)}
& \textbf{40.32} $\pm$ 0.47
& \textbf{25.48} $\pm$ 0.57
& \textbf{40.76} $\pm$ 0.44 \\

\midrule

\multirow{4}{*}{\textbf{Average}}
& Non-MoE
& 54.85 $\pm$ 0.93 & 44.06 $\pm$ 1.35 & 55.24 $\pm$ 1.09 \\
& Vanilla MoE
& 54.53 $\pm$ 0.87 & 43.60 $\pm$ 0.99 & 54.82 $\pm$ 0.78 \\
& Mean Temporal
& 55.29 $\pm$ 0.62 & 44.53 $\pm$ 0.78 & 55.62 $\pm$ 0.64 \\
& \textbf{\textsc{Trace} (Ours)}
& \textbf{55.75} $\pm$ 0.58
& \textbf{45.18} $\pm$ 0.73
& \textbf{56.10} $\pm$ 0.61 \\

\bottomrule
\end{tabular}
}
\end{table}

On average, \textsc{Trace} achieves the best balanced accuracy, surpassing the Non-MoE baseline by +0.90 percentage points, the Vanilla MoE by +1.22 percentage points, and the Mean Temporal variant by +0.46 percentage points. This leads to two observations. First, replacing the FFN with a generic token-wise MoE does not necessarily improve performance, suggesting that expert routing strategy is important for EEG modeling. Second, TemporalFormer-based temporal routing further improves over both Vanilla MoE and mean-based temporal routing, indicating that routing all channels at the same timestep with a learned cross-channel temporal context better matches the multi-channel temporal structure of EEG signals. Overall, these results show that the effectiveness of \textsc{Trace} comes from both sparse expert specialization and temporal routing.

\subsection{Model Scaling with Experts}
\label{app:model_scaling_experts}

\begin{figure}[H]
\centering

\begin{subfigure}[t]{0.49\linewidth}
    \centering
    \includegraphics[width=\linewidth]{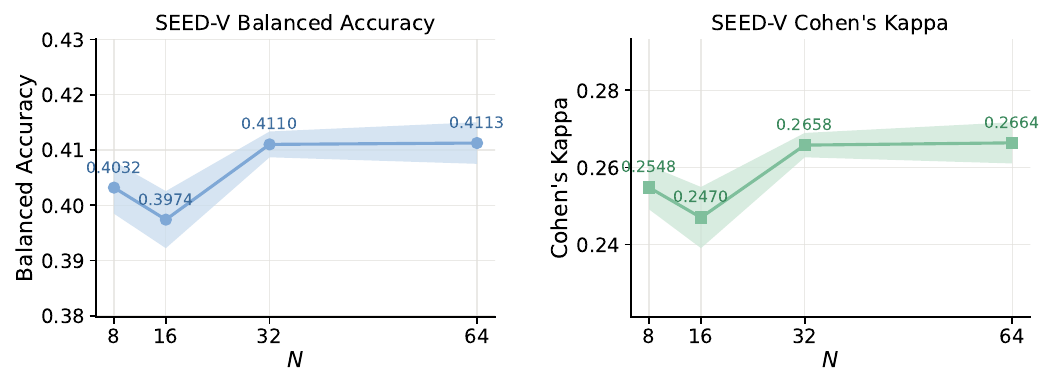}
    \caption{Number of total experts $N$ with top-$K{=}2$ fixed}
    \label{fig:n_ablation}
\end{subfigure}
\hfill
\begin{subfigure}[t]{0.49\linewidth}
    \centering
    \includegraphics[width=\linewidth]{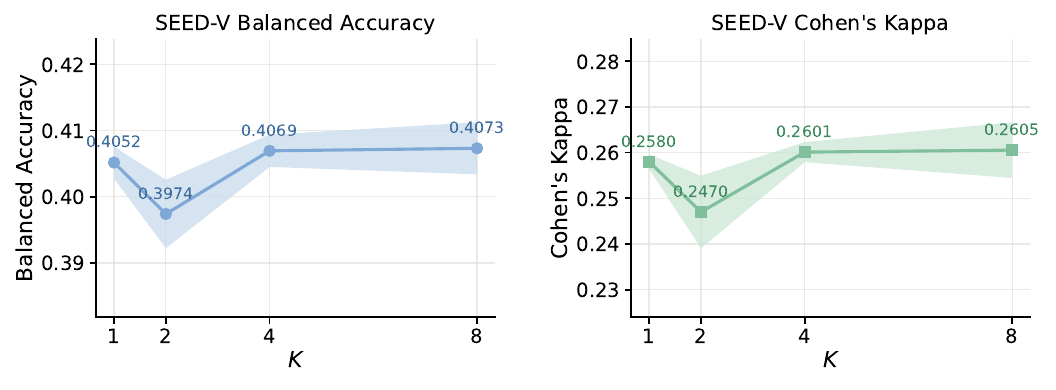}
    \caption{Number of activated experts $K$ with $N{=}16$ fixed}
    \label{fig:k_ablation}
\end{subfigure}

\caption{\textbf{Study of the Mixture-of-Experts (MoE) configuration.}
Balanced accuracy and Cohen's Kappa on SEED-V ($5$-class) when varying
(a) the total number of experts $N$ with top-$K{=}2$ fixed, and
(b) the number of activated experts $K$ with $N{=}16$ fixed.
Lines and shaded regions denote the mean and standard deviation over $5$ random seeds.}
\label{fig:nk_ablation}
\end{figure}

We analyze the impact of two key hyperparameters of CTR-FFN, \ie, the total number of specialized experts $N$ and the number of experts activated at each temporal step $K$. Figure~\ref{fig:n_ablation} shows the performance as we scale the total number of experts ($N \in \{8,16,32,64\}$) while keeping the number of activated experts fixed at $K=2$. On SEED-V~\citep{liu2021comparing}, Balanced Accuracy improves from $0.4032$ at $N=8$ to $0.4110$ at $N=32$ and then saturates, while Cohen's Kappa follows a similar trend, increasing from $0.2548$ to $0.2658$. Since training more experts increases model parameters and pre-training cost, these results suggest a trade-off between expert capacity and computational efficiency.

Figure~\ref{fig:k_ablation} presents the performance with regard to the number of top-$K$ activated experts ($K \in \{1,2,4,8\}$) from a fixed pool of $16$ total experts. For SEED-V, both Balanced Accuracy and Cohen's Kappa improve from $K=2$ to $K=4$ and then remain relatively stable, suggesting that a moderate number of activated experts is sufficient for this task. These results highlight the flexibility of top-$K$ routing: while a modest $K$ can already capture the necessary patterns for some datasets, denser expert activation does not necessarily provide further gains.

\subsection{Effect of the Shared Expert}
\label{sec:abl_shared_expert}

To evaluate the contribution of the always-on shared expert, we conduct an ablation study comparing the full \textsc{Trace} model against a variant where all experts are specialized and none are shared. All other settings, including the CTR-FFN, model architecture, pre-training corpus, and downstream fine-tuning protocol, are kept fixed.

\begin{table}[H]
\centering
\caption{Effect of the shared expert.}
\label{tab:shared_expert_ablation}
\tabstyle{0.90}{7}{1.2}{

\begin{tabular}{c|l|ccc}
\toprule
\textbf{Dataset} & \textbf{Variant} & \textbf{Balanced Acc} & \textbf{Cohen's Kappa} & \textbf{Weighted F1} \\
\midrule

\multirow{2}{*}{\textbf{FACED}}
& \textit{w/o} Shared Expert
& 62.44 $\pm$ 0.14
& 57.60 $\pm$ 0.13
& 62.77 $\pm$ 0.22 \\
& \textit{w/} Shared Expert
& \textbf{63.52} $\pm$ 0.37
& \textbf{58.83} $\pm$ 0.42
& \textbf{63.90} $\pm$ 0.43 \\

\midrule

\multirow{2}{*}{\textbf{BCIC2020}}
& \textit{w/o} Shared Expert
& 61.97 $\pm$ 0.78
& 52.46 $\pm$ 0.98
& 61.97 $\pm$ 0.82 \\
& \textit{w/} Shared Expert
& \textbf{63.15} $\pm$ 0.55
& \textbf{53.93} $\pm$ 0.69
& \textbf{63.16} $\pm$ 0.57 \\

\bottomrule
\end{tabular}
}
\end{table}

As shown in Table~\ref{tab:shared_expert_ablation}, including a shared expert provides consistent improvements on both FACED and BCIC2020 across all metrics. On average, this leads to an improvement of approximately $+1.13$ percentage points in Balanced Accuracy, $+1.35$ percentage points in Cohen's Kappa, and $+1.16$ percentage points in Weighted F1. The most notable gain is observed on BCIC2020, where Cohen's Kappa increases by nearly $1.5$ percentage points. We hypothesize that the shared expert learns to model fundamental EEG patterns that are broadly present across temporal states and channels, such as common signal properties, baseline neural activity, or general rhythmic components. The shared expert can provide a common representation pathway for general EEG patterns, while the specialized experts capture more fine-grained temporal variations. This suggests that the shared expert provides a common representation pathway for general EEG patterns, complementing the specialized experts that capture more fine-grained temporal variations.

\subsection{Effect of Patch Length}
\label{sec:abl_patch_size}

Patch length $t$ determines how raw EEG signals are grouped into temporal patches. In our setting, the signals are resampled to 200 Hz, so patch lengths of 50, 100, 200, and 400 samples correspond to temporal windows of 0.25s, 0.5s, 1s, and 2s, respectively. Each patch is encoded into one token by the time-frequency patch encoder. This token is then used as the prediction unit in multi-horizon autoregressive forecasting and as the temporal unit for expert routing. Therefore, patch length $t$ directly controls the temporal granularity of both forecasting and expert selection in \textsc{Trace}.

\begin{table}[H]
\centering
\caption{Ablation on patch length $t$.}
\label{tab:patch_size_ablation}
\tabstyle{0.88}{6}{1.2}{

\begin{tabular}{l|c|ccc}
\toprule
\textbf{Dataset} & \textbf{Patch length $t$} & \textbf{Cohen's Kappa} & \textbf{Weighted F1} & \textbf{Balanced Accuracy} \\
\midrule

\multirow{4}{*}{\textbf{FACED}}
& 50
& 52.80 $\pm$ 1.88
& 58.22 $\pm$ 1.64
& 58.36 $\pm$ 1.70 \\
& 100
& 53.23 $\pm$ 1.49
& 58.56 $\pm$ 1.29
& 58.74 $\pm$ 1.35 \\
& \textbf{200}
& \textbf{58.83} $\pm$ 0.42
& \textbf{63.90} $\pm$ 0.43
& \textbf{63.52} $\pm$ 0.37 \\
& 400
& 53.50 $\pm$ 0.83
& 59.13 $\pm$ 0.64
& 58.82 $\pm$ 0.73 \\

\midrule

\multirow{4}{*}{\textbf{ISRUC}}
& 50
& 75.37 $\pm$ 0.72
& 80.65 $\pm$ 0.55
& 78.77 $\pm$ 0.15 \\
& 100
& 73.44 $\pm$ 2.86
& 78.87 $\pm$ 2.63
& 78.26 $\pm$ 1.12 \\
& \textbf{200}
& \textbf{75.41} $\pm$ 0.29
& \textbf{80.77} $\pm$ 0.18
& \textbf{79.41} $\pm$ 0.58 \\
& 400
& 75.06 $\pm$ 1.12
& 80.57 $\pm$ 0.95
& 78.77 $\pm$ 0.56 \\

\bottomrule
\end{tabular}
}
\end{table}

We compare four patch lengths $t$ while keeping the model architecture, pre-training corpus, and downstream fine-tuning protocol fixed. As shown in Table~\ref{tab:patch_size_ablation}, the 1s patch achieves the best performance on both FACED and ISRUC across all three metrics. The effect is especially clear on FACED, where the 1s patch improves Balanced Accuracy by nearly 4.7 percentage points over the neighboring settings. ISRUC shows a smaller but consistent gain, suggesting that the 1s patch also transfers well to longer sleep-stage epochs.

\textbf{Compatibility with downstream windows.}
A 1s patch naturally matches common EEG window lengths used in downstream benchmarks. For example, SEED-V uses 1s clips, PhysioNet-MI uses 4s windows, and ISRUC uses 30s epochs. With a 1s patch, these inputs can be represented as an integer number of tokens without unnecessary padding or truncation. This choice is also consistent with recent EEG foundation models, where 1s EEG segments are commonly used as the basic modeling unit~\citep{jiang2024large,wang2025cbramod,ma2025codebrain}. Smaller patches increase the number of temporal tokens and the routing cost of the CTR-FFN, while larger patches can be less suitable for short-window datasets.

\textbf{Preserving waveform structure.}
A 1s patch also better preserves physiologically meaningful EEG patterns. Many EEG events occur at sub-second to second-level scales, such as K-complexes around 1s, sleep spindles lasting 0.5--2s, and event-related potentials such as P300 around 0.3--0.6s. Very small patches may split these patterns across multiple tokens, while very large patches may merge fast transients into a coarse representation. The 1s setting therefore provides a practical balance between local waveform preservation, temporal abstraction, and computational efficiency.

\subsection{Embedding Design Ablation}
We further ablate the embedding design of \textsc{Trace}. All variants keep the backbone, pre-training objective, and downstream fine-tuning protocol fixed, and only modify the patch encoder. The results are summarized in Tables~\ref{tab:branch_design_ablation}--\ref{tab:cnn_kernel_ablation}.

\paragraph{Branch design.}
We first examine whether both the time-domain CNN branch and the spectral FFT branch are necessary. Removing the time branch causes a substantial drop, decreasing Balanced Accuracy by $23.25$ percentage points on FACED and $6.03$ points on BCIC2020-3. In contrast, removing the FFT branch only mildly affects performance, with drops of $0.40$ and $2.11$ points on the two datasets. This suggests that the time-domain CNN branch carries the dominant discriminative signal, while the FFT branch provides complementary spectral cues for tasks with stronger frequency-band structure.

\begin{table}[H]
\centering
\caption{{Anlysis of time-frequency branch design.}}
\label{tab:branch_design_ablation}
\tabstyle{0.82}{7}{1.2}{
\begin{tabular}{l|cc|cc}
\toprule
\multirow{2}{*}{\textbf{Branch Design}}
& \multicolumn{2}{c|}{\textbf{FACED}}
& \multicolumn{2}{c}{\textbf{BCIC2020-3}} \\
\cmidrule(lr){2-3}\cmidrule(lr){4-5}
& \textbf{Balanced Acc} & \textbf{Kappa}
& \textbf{Balanced Acc} & \textbf{Kappa} \\
\midrule

Full dual-branch
& \textbf{0.6352} $\pm$ 0.0037
& \textbf{0.5883}
& \textbf{0.6315} $\pm$ 0.0055
& \textbf{0.5393} \\

Time-only (\textit{w/o} FFT)
& 0.6312 $\pm$ 0.0083
& 0.5835
& 0.6104 $\pm$ 0.0151
& 0.5130 \\

FFT-only (\textit{w/o} Time)
& 0.4027 $\pm$ 0.0129
& 0.3250
& 0.5712 $\pm$ 0.0214
& 0.4640 \\

\bottomrule
\end{tabular}
}
\end{table}

\paragraph{Fusion strategy.}
We examine whether the learnable sigmoid gate is necessary for fusing the time-domain and FFT-based branches. Compared with the default sigmoid-gated fusion, fixed summation decreases Balanced Accuracy by $1.36$ percentage points on FACED and $0.99$ points on BCIC2020-3. It also leads to a larger standard deviation on BCIC2020-3. This suggests that the two branches should not be combined with a fixed equal-weight rule; the learnable gate provides an adaptive fusion mechanism that dynamically balances temporal and spectral information.

\begin{table}[H]
\centering
\caption{Analysis of branch fusion strategy.}
\label{tab:fusion_ablation}
\tabstyle{0.80}{7}{1.2}{
\begin{tabular}{l|cc|cc}
\toprule
\multirow{2}{*}{\textbf{Fusion Design}}
& \multicolumn{2}{c|}{\textbf{FACED}}
& \multicolumn{2}{c}{\textbf{BCIC2020-3}} \\
\cmidrule(lr){2-3}\cmidrule(lr){4-5}
& \textbf{Balanced Acc} & \textbf{Kappa}
& \textbf{Balanced Acc} & \textbf{Kappa} \\
\midrule

Sigmoid-gated fusion
& \textbf{0.6352} $\pm$ 0.0037
& \textbf{0.5883}
& \textbf{0.6315} $\pm$ 0.0055
& \textbf{0.5393} \\

Fixed sum fusion
& 0.6216 $\pm$ 0.0118
& 0.5731
& 0.6216 $\pm$ 0.0216
& 0.5270 \\

\bottomrule
\end{tabular}
}
\end{table}

\paragraph{CNN kernel scale.}
We further examine the effect of temporal kernel scale in the time-domain CNN branch. As shown in Table~\ref{tab:cnn_kernel_ablation}, both single-kernel variants underperform the multi-scale default. On FACED, using only the narrow kernel $q=25$ and the wide kernel $q=99$ decreases Balanced Accuracy by $1.56$ and $2.09$ percentage points, respectively. On BCIC2020-3, the gap is smaller: $q=25$ trails the default by $0.88$ points, while $q=99$ is only $0.27$ points lower.

These results suggest that informative EEG waveform patterns appear at multiple temporal scales. A single narrow or wide kernel can capture part of the signal, but the multi-scale design $\mathcal{Q}_{\mathrm{temp}}=\{25,49,99\}$ provides a more robust time-domain embedding across tasks with different temporal dynamics.

\begin{table}[H]
\centering
\caption{Ablation on temporal CNN kernel scale.}
\label{tab:cnn_kernel_ablation}
\tabstyle{0.90}{7}{1.2}{
\begin{tabular}{l|cc|cc}
\toprule
\multirow{2}{*}{\textbf{Temporal Kernel Set}}
& \multicolumn{2}{c|}{\textbf{FACED}}
& \multicolumn{2}{c}{\textbf{BCIC2020-3}} \\
\cmidrule(lr){2-3}\cmidrule(lr){4-5}
& \textbf{Balanced Acc} & \textbf{Kappa}
& \textbf{Balanced Acc} & \textbf{Kappa} \\
\midrule

Multi-scale $\mathcal{Q}_{\mathrm{temp}}=\{25,49,99\}$
& \textbf{0.6352} $\pm$ 0.0037
& \textbf{0.5883}
& \textbf{0.6315} $\pm$ 0.0055
& \textbf{0.5393} \\

Single-scale $\mathcal{Q}_{\mathrm{temp}}=\{25\}$
& 0.6196 $\pm$ 0.0089
& 0.5697
& 0.6227 $\pm$ 0.0206
& 0.5283 \\

Single-scale $\mathcal{Q}_{\mathrm{temp}}=\{99\}$
& 0.6143 $\pm$ 0.0140
& 0.5641
& 0.6288 $\pm$ 0.0137
& 0.5360 \\

\bottomrule
\end{tabular}
}
\end{table}

\subsection{Effect of Auxiliary Expert-Balancing Loss}
\label{app:aux_routing_loss}

We examine whether the auxiliary expert-balancing loss $\mathcal{L}_{\mathrm{aux}}$ is necessary for maintaining diverse expert routing in CTR-FFN. This analysis uses the ablation setting with $N{=}16$, $K{=}4$, and TemporalFormer routing. We compare the default model in this setting with an ablated variant trained without the auxiliary expert-balancing loss, and analyze the routed experts on four downstream datasets: FACED, SEED-V, BCIC2020-3, and PhysioNet-MI. To quantify cross-dataset expert overlap, we compute the Jaccard similarity between the top-4 routed expert sets of different datasets. For two expert sets $A$ and $B$, it is defined as $J(A,B)=|A\cap B|/|A\cup B|$. A higher Jaccard similarity indicates that different datasets rely on more similar experts, suggesting more concentrated and less task-specific routing.

\begin{table}[H]
\centering
\caption{Top-$4$ routed experts with and without auxiliary expert-balancing loss.}
\label{tab:top4_expert_overlap}
\tabstyle{0.55}{8}{1.2}{
\begin{tabular}{lcc}
\toprule
\textbf{Dataset} & \textbf{Default} & \textbf{\textit{w/o} Aux. Loss} \\
\midrule
FACED & $\{6,3,5,15\}$ & $\{12,\mathbf{9},6,2\}$ \\
SEED-V & $\{13,16,1,8\}$ & $\{4,\mathbf{9},\underline{8},5\}$ \\
BCIC2020-3 & $\{2,13,11,8\}$ & $\{\mathbf{9},\underline{8},12,10\}$ \\
PhysioNet-MI & $\{13,6,5,3\}$ & $\{\mathbf{9},\underline{8},7,13\}$ \\
\bottomrule
\end{tabular}
}
\vspace{3pt}
\parbox{0.60\linewidth}{
\footnotesize \emph{Note:} Expert~\textbf{9} appears in the top-$4$ experts for all four datasets under the ablated setting; expert~\underline{$8$} appears in three out of four datasets.
}
\end{table}

\begin{table}[H]
\centering
\caption{Quantitative evidence of routing concentration.}
\label{tab:routing_jaccard}
\tabstyle{0.95}{7}{1.2}{
\begin{tabular}{lccc}
\toprule
\textbf{Metric} & \textbf{Default} & \textbf{w/o Aux. Loss} & \textbf{Change} \\
\midrule
Mean pairwise Jaccard of top-4 experts & $0.203$ & $0.314$ & $+55\%$ \\
Universal experts in top-4 across 4 datasets & $0$ & $1$ & $+1$ \\
Experts appearing in $\geq 3$ datasets' top-4 & $1$ & $2$ & $+1$ \\
Task-specific experts appearing in only 1 dataset & $5$ & $5$ & $0$ \\
\bottomrule
\end{tabular}
}
\end{table}

\textbf{1) Removing the auxiliary loss increases cross-dataset expert overlap.}
As shown in Table~\ref{tab:routing_jaccard}, the mean pairwise Jaccard similarity rises from $0.203$ to $0.314$ after removing the auxiliary expert-balancing loss, corresponding to a relative increase of $55\%$. This indicates that the dominant experts selected by different downstream datasets become more similar without the balancing constraint.

\textbf{2) Removing the auxiliary loss creates a universal dominant expert.}
As shown in Table~\ref{tab:top4_expert_overlap}, in the default model, no expert appears in the top-4 routed experts for all four datasets. In contrast, without the auxiliary expert-balancing loss, expert~$9$ appears in the top-4 routed experts for FACED, SEED-V, BCIC2020-3, and PhysioNet-MI, while expert~$8$ appears in three out of four datasets. This suggests that the router becomes more dependent on a small subset of experts when the auxiliary loss is removed. Overall, these results indicate that the auxiliary expert-balancing loss helps prevent routing collapse and encourages more diverse expert utilization across downstream tasks.

\begin{figure}[t]
\centering
\includegraphics[width=\linewidth]{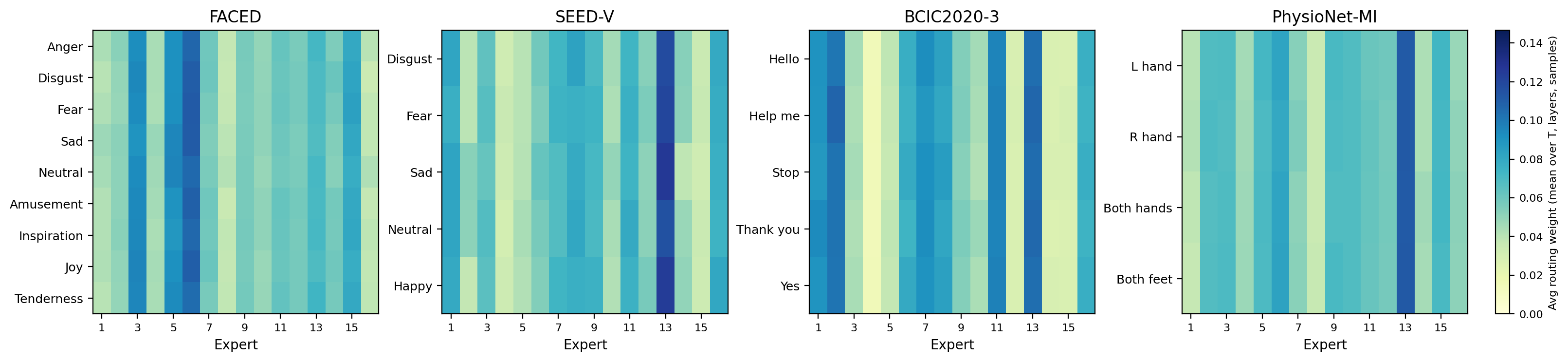}\\[2pt]
\includegraphics[width=\linewidth]{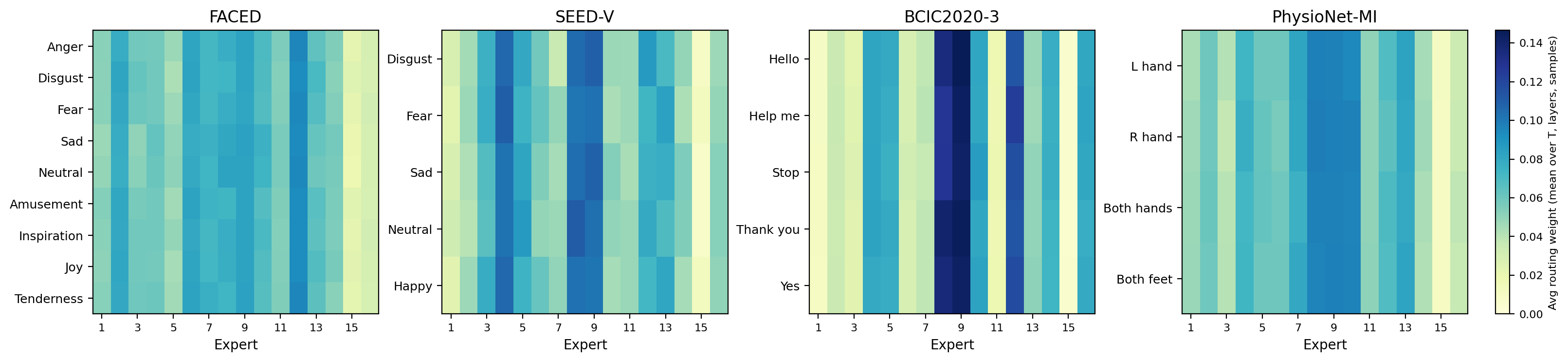}
\caption{\textbf{Effect of auxiliary expert-balancing loss on per-class routing.}
Average routing weights over time, layers, and samples for four downstream datasets under the auxiliary-loss ablation setting with $N{=}16$, $K{=}4$, and TemporalFormer routing. \textit{Top:} default model with auxiliary expert-balancing loss. \textit{Bottom:} ablated model without auxiliary expert-balancing loss. Removing the auxiliary loss leads to more concentrated expert usage: expert~$9$ appears in the top-$4$ experts for all datasets, and the mean pairwise top-$4$ Jaccard similarity increases from $0.203$ to $0.314$ ($+55\%$).}
\label{fig:aux_loss}
\end{figure}

Figure~\ref{fig:aux_loss} provides a qualitative visualization of the same trend, where the ablated model shows more concentrated routing weights across datasets.

\end{document}